\documentclass[journal]{IEEEtran}
\usepackage{amsmath,amssymb} % define this before the line numbering.
\usepackage{tikz}
\usepackage{graphicx}
\usepackage{comment}
\usepackage{color}

\usepackage{hyperref}
\usepackage{epsfig}
\usepackage[ruled,norelsize]{algorithm2e}
\usepackage[normalem]{ulem}
\usepackage{subfigure}
\usepackage{mathtools}
\usepackage{wrapfig}

\def\onedot{. }
\def\eg{\emph{e.g.},} 
\def\ie{\emph{i.e.},}

\def\etal{\emph{et al}\onedot}

\newcommand{\R}{\mathbb{R}}
\def\eg{\emph{e.g.},} 

\newcommand{\Tref}[1]{Table~\ref{#1}}
\newcommand{\Eref}[1]{Eq.~(\ref{#1})}
\newcommand{\Fref}[1]{Fig.~\ref{#1}}
\newcommand{\Aref}[1]{Alg.~\ref{#1}}

\newcommand{\Cref}[1]{Chap.~\ref{#1}}
\newcommand{\Sref}[1]{Sec.~\ref{#1}}
\ifCLASSINFOpdf
\else
\fi
\begin{document}
%
% paper title
% Titles are generally capitalized except for words such as a, an, and, as,
% at, but, by, for, in, nor, of, on, or, the, to and up, which are usually
% not capitalized unless they are the first or last word of the title.
% Linebreaks \\ can be used within to get better formatting as desired.
% Do not put math or special symbols in the title.
\title{ACP++: Action Co-occurrence Priors \\for Human-Object Interaction Detection}
%
%
% author names and IEEE memberships
% note positions of commas and nonbreaking spaces ( ~ ) LaTeX will not break
% a structure at a ~ so this keeps an author's name from being broken across
% two lines.
% use \thanks{} to gain access to the first footnote area
% a separate \thanks must be used for each paragraph as LaTeX2e's \thanks
% was not built to handle multiple paragraphs
%

\author{Dong-Jin Kim,~\IEEEmembership{Member,~IEEE,}
        Xiao Sun,~\IEEEmembership{Member,~IEEE,}\\
        Jinsoo Choi,~\IEEEmembership{Member,~IEEE,}
        Stephen Lin,~\IEEEmembership{Member,~IEEE,}
        In So Kweon,~\IEEEmembership{Member,~IEEE,}% <-this % stops a space
\IEEEcompsocitemizethanks{\IEEEcompsocthanksitem D. Kim, J. Choi, and I. S. Kweon are with the School of Electrical and
Computer Engineering, KAIST, Daejeon, Republic of Korea.\protect\\
E-mail: \{djnjusa,jinsc37,iskweon77\}@kaist.ac.kr
\IEEEcompsocthanksitem X. Sun and S. Lin are with the Visual Computing group, Microsoft Research, Beijing, China.
\protect\\
E-mail: \{xias,stevelin\}@microsoft.com
}% <-this % stops an unwanted space
\thanks{}}

% note the % following the last \IEEEmembership and also \thanks - 
% these prevent an unwanted space from occurring between the last author name
% and the end of the author line. i.e., if you had this:
% 
% \author{....lastname \thanks{...} \thanks{...} }
%                     ^------------^------------^----Do not want these spaces!
%
% a space would be appended to the last name and could cause every name on that
% line to be shifted left slightly. This is one of those "LaTeX things". For
% instance, "\textbf{A} \textbf{B}" will typeset as "A B" not "AB". To get
% "AB" then you have to do: "\textbf{A}\textbf{B}"
% \thanks is no different in this regard, so shield the last } of each \thanks
% that ends a line with a % and do not let a space in before the next \thanks.
% Spaces after \IEEEmembership other than the last one are OK (and needed) as
% you are supposed to have spaces between the names. For what it is worth,
% this is a minor point as most people would not even notice if the said evil
% space somehow managed to creep in.

% The paper headers
\markboth{Journal of \LaTeX\ Class Files,~Vol.~14, No.~8, August~2015}%
{Shell \MakeLowercase{\textit{et al.}}: Bare Demo of IEEEtran.cls for IEEE Journals}
% The only time the second header will appear is for the odd numbered pages
% after the title page when using the twoside option.
% 
% *** Note that you probably will NOT want to include the author's ***
% *** name in the headers of peer review papers.                   ***
% You can use \ifCLASSOPTIONpeerreview for conditional compilation here if
% you desire.

% make the title area
\maketitle

% As a general rule, do not put math, special symbols or citations
% in the abstract or keywords.
\begin{abstract}
A common problem in the task of human-object interaction (HOI) detection is that numerous HOI classes have only a small number of labeled examples, resulting in training sets with a long-tailed distribution. 
The lack of positive labels can lead to low classification accuracy for these classes. 
Towards addressing this issue, we observe that there exist natural correlations and anti-correlations among human-object interactions. 
In this paper, we model the correlations as {\em action co-occurrence matrices} and present techniques to learn these priors and leverage them for more effective training, especially on rare classes.
The efficacy of our approach is demonstrated experimentally, where the performance of our approach consistently improves over
the state-of-the-art methods on both of the two leading HOI detection benchmark datasets, HICO-Det and V-COCO.
\end{abstract}

% Note that keywords are not normally used for peerreview papers.

\IEEEpeerreviewmaketitle

%%%%%%%%% BODY TEXT
%%%%%%%%%----1----%%%%%%%%%%%%%%%%%%%%%%%%%%%%%%%%%%%%%%%%%%%%%%%%%%%%%%%%%%%%%%%%%%%%%%%%%%%%%
\section{Introduction}
\IEEEPARstart{H}{uman}-object interaction (HOI) detection aims to localize humans and objects in an image and infer the relationships between them. 
An HOI is typically represented as a human-action-object triplet with the corresponding bounding boxes and classes. 
Detecting these interactions is a fundamental challenge in visual recognition that requires both an understanding of object information and high-level knowledge of interactions.

\begin{figure}[t]
	\centering
	\includegraphics[width=1\linewidth,keepaspectratio]{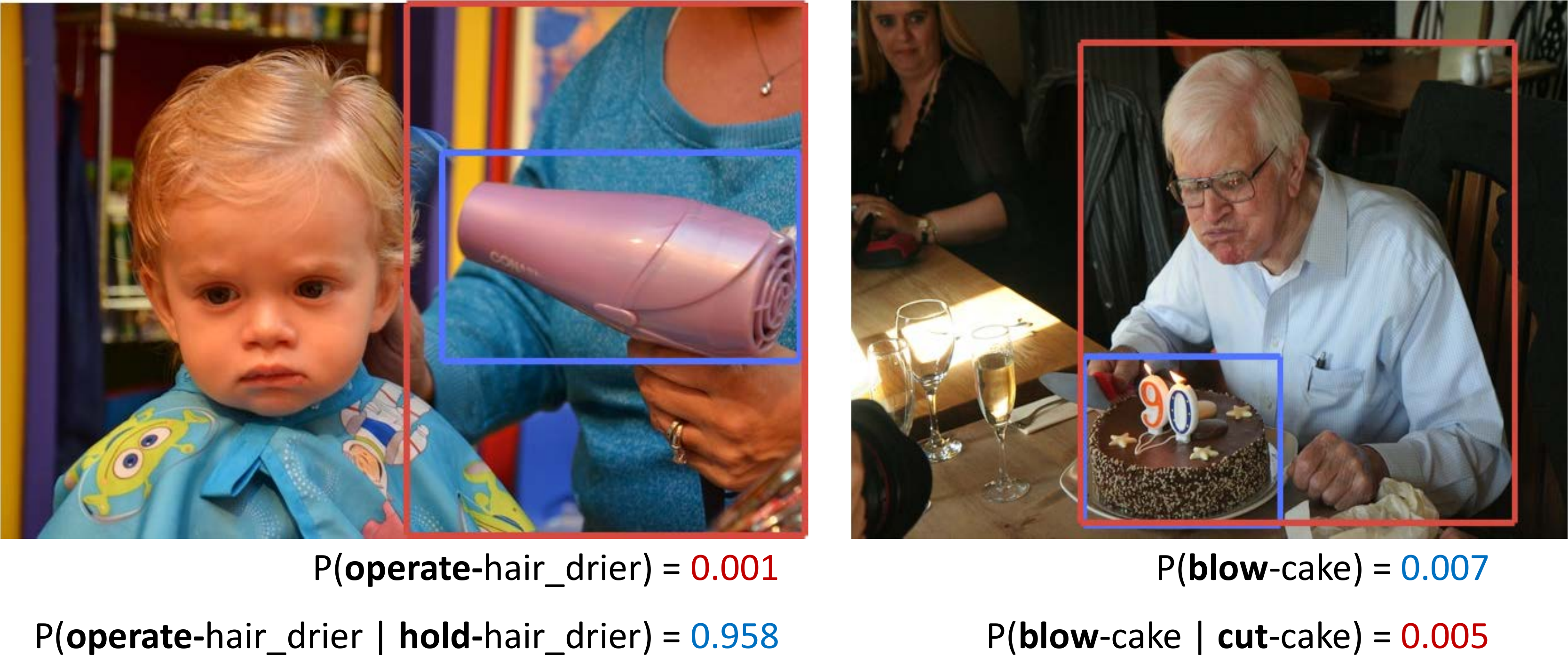}
	\vspace{-2mm}
	\caption{Examples of action co-occurrence in HOI detection datasets. 
	The marginal/conditional probability values are computed from the distribution of training labels. Intuitively, detection of rarely labeled HOIs (operate-hair dryer) can be facilitated by detection of commonly co-occurring HOIs (hold-hair dryer). Also, non-detection of rare HOIs (blow-cake) can be aided by detection of incompatible HOIs (cut-cake). 
	We leverage this intuition as a prior to learn an HOI detector effective on long-tailed datasets.
	}
	\vspace{-2mm}
	\label{fig:teaser}
\end{figure}

A major issue that exists in HOI detection is that its datasets suffer from long-tailed distributions, in which many HOI triplets have few labeled instances.
Similar to datasets for general visual relationship detection (VRD)~\cite{lu2016visual}, a reason for this is missing labels, where the annotation covers only a subset of the interactions present in an image.
For the widely-used HICO-Det dataset~\cite{chao2018learning}, 462 out of the 600 HOI classes have fewer than 10 training samples. 
For such classes, the lack of positive labels can lead to inadequate training and low classification performance.
In particular, since the supervision for the layer weights is mostly 0 for rare classes, confusion occurs for these classes.
{How to alleviate performance degradation on rare classes is thus a key issue in HOI detection.}
%As a result, the weight parameters affecting the classification of rare classes will not receive sufficient gradients from positive labels during training. 

To address the problem of long-tailed distributions, we propose to take advantage of natural \emph{co-occurrences} in human actions. 
In other words, given a pair of a human and an object, multiple actions or interactions can happen at the same time.
For example, the HOI of `operate-hair dryer' is rarely labeled and consequently hard to detect in the left image of \Fref{fig:teaser}. 
However, `operate-hair dryer' often occurs when the more commonly labeled HOI of `hold-hair dryer' is present. As a result, detection of `operate-hair dryer' can be facilitated by detection of `hold-hair dryer' in an image. On the other hand, the detection of an HOI may preclude other incompatible HOIs, such as for `cut-cake' and `blow-cake' in the right image of \Fref{fig:teaser}.

In this paper, we introduce {the new concept of utilizing co-occurring actions as prior knowledge, termed as} action co-occurrence priors (ACPs), to train an HOI detector. 
In particular, we count the co-occurrences of class labels present in the training data and leverage this knowledge to effectively train our model.
{In contrast to language-based prior knowledge which requires external data sources~\cite{kim2019image,lu2016visual,yu2017visual}, co-occurrence priors can be easily obtained from the label statistics of the target dataset.}
{We also propose} two novel ways to exploit them. 
First, we {design} a {neural network with hierarchical structure} where the classification is initially performed with respect to {\em action groups}.
{Each action group is defined by one anchor action, where the anchor actions are mutually exclusive according to the co-occurrence prior.}
{Then, our model predicts the fine-grained HOI class within the action group.}
With this approach, higher performance is attained by learning dedicated classifiers for distinguishing closely-related actions within a group.
%\steve{This hierarchical architecture also incorporates a self-attention module which enriches the semantic meaning of features based on surrounding human-object pairs.}
%\djkim{In order to explicitly augment semantic meanings of surrounding human-object pairs, we additionally add a self-attention module motivated by non-local networks~\cite{wang2018non} upon our hierarchical architecture.}
{Second, we present a technique that employs knowledge distillation~\cite{hinton2015distilling} to expand HOI labels 
so they can have more positive labels for potentially co-occurring actions.} 
%\steve{Along with this knowledge distillation, linguistic prior knowledge via a word embedding is introduced to further alleviate dataset bias.}
During training, the predictions are regularized by the refined objectives to improve robustness, especially for rare classes.
%in the data distribution tail.
%\djkim{In order to further alleviate the dataset bias by exploit linguistic prior knowledge, we additionally introduce a word embedding loss along with the knowledge distillation loss.}
To the best of our knowledge, this is the first work to leverage label co-occurrences in HOI detection to alleviate the long-tailed distribution problem.

%(Our solution is....)
%In this paper, given a co-occurrence matrix constructed from the label distribution of the dataset, we introduce two different ways to exploit the co-occurrence prior. 
%First, given this co-occurrence information, we can cluster the HOI classes into several action groups. 
%We use this group label as an auxiliary ground truth label and propose a hierarchical architecture effectively learning the group prior. 
%Second, we refine the original ground truth HOI label with the co-occurrence matrix to address the problem of the long-tailed distribution of the original label. 

%(Summary of our contributions)
{The main contributions of this work can be summarized as: 
(1) The novel concept of explicitly leveraging correlations among HOI labels to address the problem of long-tailed distributions in HOI detection; 
(2) Two orthogonal ways to leverage action co-occurrence priors, namely through a proposed hierarchical architecture and HOI label expansion via knowledge distillation.
The resulting model is shown to be consistently advantageous in relation to state-of-the-art techniques on both the HICO-Det~\cite{chao2018learning} and V-COCO~\cite{gupta2015visual} benchmark datasets.} 
%The main contributions of this work are summarized as follows. 
%(1) The concept of explicitly leveraging correlations among HOI labels to address the problem of long-tailed distributions in HOI detection. 
%(2) Two orthogonal ways to leverage action co-occurrence priors, namely through a proposed hierarchical architecture and knowledge distillation for training. 
%(3) The resulting model which is shown to be consistently advantageous in relation to state-of-the-art techniques on both HICO-Det~\cite{chao2018learning} and V-COCO~\cite{gupta2015visual} benchmark datasets. 

%\steve{Based on the reviewer's comment, I think it is fine to motivate the new components in the next paragraph, rather than discuss them two paragraphs earlier, where it hurts the presentation flow a bit. Please check if what I've written below is consistent with your understanding.}\djkim{(checked! I think this paragraph would be perfect in terms of  the presentation flow.)}\steve{If there is any concern about the content, specifically for the word embedding loss, please let me know. I'm not sure how correct my understanding is.}

This work is an extension of our previous conference paper~\cite{kim2020detecting}. The primary differences are as follows: 
(1) We extend our architecture to incorporate a self-attention module, whose purpose is to enrich the semantic content of each human-object pair by accounting for surrounding human-object pairs. In this way, global image context is leveraged to better understand the interaction of a human and object.
%(1) We extend our architecture by adding a self-attention module motivated by non-local networks~\cite{wang2018non} to explicitly augment semantic meanings of surrounding human-object pairs. 
(2) We additionally exploit linguistic prior knowledge through a word embedding regression loss for object classes. In the word embedding, object categories that are more semantically similar are closer together. With this loss, learning for rare HOI labels can benefit from more common HOI labels that are semantically similar, further alleviating dataset bias.
%(2) Also, we added a word embedding regression loss for object classes upon the knowledge distillation loss to exploit linguistic prior knowledge.
This extended method including both of the new components is called ACP++. We show that ACP++ outperforms our previous ACP in all the application scenarios we demonstrate.
In addition, we expand our experimental results and analysis to show multiple aspects of our proposed method's algorithmic behavior.

%%%%%%%%%----2----%%%%%%%%%%%%%%%%%%%%%%%%%%%%%%%%%%%%%%%%%%%%%%%%%%%%%%%%%%%%%%%%%%%%%%%%%%%%%
\section{Related Work}

The HOI detection task is rooted in visual relationship detection (VRD) and %is similar to the task of 
scene graph generation.
In this section, we summarize the recent work on HOI detection, VRD, and scene graph generation.
We also review works that utilize a label hierarchy in multi-label learning.

\noindent\textbf{Human-Object Interaction} 
Human-Object Interaction was originally studied in the context of recognizing the function or `affordance' of objects~\cite{delaitre2012scene,gibson2014ecological,grabner2011makes,gupta2007objects,stark1991achieving}. Early works focus on learning more discriminative features combined with variants of SVM classifiers~\cite{delaitre2010recognizing,delaitre2011learning,yao2010grouplet}, and leverage the relationship with human poses for better representation~\cite{delaitre2010recognizing,delaitre2011learning,yao2011human} or mutual context modeling~\cite{yao2010modeling}.

Recently, a completely data-driven approach based on convolutional neural networks (CNNs) has brought dramatic progress to HOI. Many of the pioneering works constructed large-scale image datasets~\cite{chao2018learning,chao2015hico,gupta2015visual,zhuang2018hcvrd} 
to set new benchmarks in this field. 
Since then, significant progress has been achieved in using CNNs for this problem~\cite{bansal2020detecting,chao2018learning,gao2018ican,gao2020drg,gkioxari2018detecting,gupta2019no,hou2020visual,kato2018compositional,li2020detailed,li2019transferable,liao2020ppdm,qi2018learning,shen2018scaling,ulutan2020vsgnet,wan2019pose,wang2020learning,xu2019interact,xu2019learning,zhong2020polysemy}. 

Most of these works follow a two-step scheme of CNN feature extraction and multi-information fusion, where the multiple information may include 
human and object appearance~\cite{chao2018learning,gao2018ican,gkioxari2018detecting,qi2018learning}; 
box relation (either box configuration or spatial map)~\cite{bansal2020detecting,chao2018learning,gkioxari2018detecting,gupta2019no,xu2019learning};
object category~\cite{bansal2020detecting,gao2020drg,gupta2019no,peyre2019detecting}; 
human pose~\cite{gupta2019no,li2020detailed,li2019transferable};
%object location, category or appearance~\cite{chao2018learning,gkioxari2018detecting, gao2018ican, xu2019interact, gupta2019no, fang2018pairwise, wan2019pose}; human location, pose or appearance~\cite{gao2018ican, xu2019interact, mallya2016learning, gupta2019no, fang2018pairwise, wan2019pose}; 
and particularly, linguistic prior knowledge~\cite{gao2020drg,kato2018compositional,peyre2019detecting}. 
More recent works tend to combine these various cues~\cite{gupta2019no,li2019transferable,wan2019pose,xu2019interact}.
%human appearance and/or object appearance~\cite{chao2018learning,gkioxari2018detecting,qi2018learning,gao2018ican,xu2019learning,gupta2019no,li2019transferable,wang2019deep,peyre2019detecting,wan2019pose,zhou2019relation}
%box relation (including geometric configuration)~\cite{chao2018learning,gkioxari2018detecting,gao2018ican,xu2019learning,gupta2019no,li2019transferable,wang2019deep,wan2019pose}
%object label~\cite{gupta2019no}
%human pose~\cite{gupta2019no,li2019transferable,wan2019pose,xu2019interact}
% ~\cite{kato2018compositional, bansal2019detecting, peyre2019detecting, xu2019learning, li2019transferable, yu2017visual}
These works differ from one another mainly in their techniques for exploiting external knowledge priors. 
Kato~\etal~\cite{kato2018compositional} incorporate information from WordNet~\cite{miller1995wordnet} using a Graph Convolutional Network (GCN)~\cite{kipf2016semi} and learn to compose new HOIs. 
Xu~\etal~\cite{xu2019learning} also use a GCN to model the general dependencies among actions and object categories by leveraging a VRD dataset~\cite{lu2016visual}. 
%Bansal~\etal~\cite{bansal2019detecting} use the off-the-shelf word2vec~\cite{mikolov2013distributed} representation for capturing functional similarities between objects. 
%Gkioxari~\etal~\cite{gkioxari2018detecting} proposed a human-centric multi-task learning approach by modifying the Faster R-CNN~\cite{ren2015faster}.
%Qi~\etal~\cite{qi2018learning} introduce a Graph Parsing Neural Network (GPNN) by repesenting the HOI structures in graphs in a graph representation.
%Gao~\etal~\cite{gao2018ican} propose instance-centric attention model to attend on salient regions of image for detecting HOIs.
Li~\etal~\cite{li2019transferable} utilize interactiveness knowledge learned across multiple HOI datasets. 
Peyre~\etal~\cite{peyre2019detecting} transfer knowledge from triplets seen at training to new unseen triplets at test time by analogy reasoning.

%Wan~\etal~\cite{wan2019pose} ----.
%Wang~\etal~\cite{wang2019deep} ----.
%Zhou~\etal~\cite{zhou2019relation} ----.

%Yu et al.~\cite{yu2017visual} leverage the strong correlations between the predicate and the $(subj, obj)$ pair to predict predicates conditioned on the subjects and the objects using Linguistic Knowledge Distillation~\cite{hinton2015distilling}.

% Also, some argue that predicting a verb instead of an HOI label is effective in the case of zero-shot learning~\cite{shen2018scaling, gupta2019no, bansal2019detecting}. 

Different from the previous works that focus on network architecture and human representation, we propose an orthogonal perspective to reformulate the target action label space and corresponding loss function by leveraging \emph{co-occurrence} relationships among action classes for HOI detection. 
To the best of our knowledge, this is the first work that leverages the co-occurrence relationship between actions for HOI recognition.
In principle, our method is complementary to all of the previous works and can be combined with any of them. 
%For our experiments, we implemented our approach on a baseline presented in~\cite{gupta2019no}, with details given in \Sref{sec.architecture}.
We have implemented our approach on several existing HOI detection architectures~\cite{gao2020drg,gupta2019no,liao2020ppdm}, and we mainly evaluated our method with a baseline presented in~\cite{gupta2019no} with details described in \Sref{sec.architecture}.

\noindent\textbf{Visual Relationship Detection and Scene Graph Generation} 
The closest problems to HOI detection are Visual Relationship Detection (VRD)~\cite{dai2017detecting,li2017vip,plummer2016phrase,yang2018shuffle,yin2018zoom,yu2017visual,zhan2019exploring,zhang2017visual,zhang2017relationship,zhuang2017towards} and scene graph generation~\cite{li2018factorizable,gu2019scene,li2017scene,woo2018linknet,qi2019attentive,wang2019exploring,xu2017scene,yang2018graph,zellers2018neural}, which deal with general visual relationships between two arbitrary objects. 
In VRD and scene graph datasets~\cite{krishna2017visual,lu2016visual}, the types of visual relationships that are modeled include verb (action), preposition, spatial and comparative phrase. 
Scene graph generation, VRD, and HOI detection share common challenges such as long-tailed distributions and even zero-shot problems~\cite{lu2016visual}. 
In this paper, we focus on HOI detection, as co-occurrences of human-object interactions are often strong, but the proposed technique could be extended to model the general co-occurrences that exist in visual relationships.
%While recent scene graph generation models also represent co-occurrence implicitly via architecture,
While several other works on different tasks explore
co-occurrence or contextual relation implicitly~\cite{galleguillos2008object,ladicky2010graph,ladicky2013inference,li2011superpixel,mottaghi2014role,sulimowicz2018superpixel,sutton2006introduction,vo2014modeling,yao2012describing,yao2018exploring},
 we explicitly model the co-occurrence of the data so that we can more efficiently exploit co-occurrence priors. In addition, as mentioned before, our method is complementary to the existing scene graph generation methods.

% Similar to HOI detection, visual relationship detection (VRD) is a task to predict subjects, objects, and predicates.
% The difference is that VRD requires to correctly predict a subject class, without being limited to `person.’
% Thus, the HOI task can be regarded as a special case of VRD in that it restricts the subject to `person.'

% As Lu~\etal first formalizes the VRD task, they first raise a long-tailed distribution of the relationship understanding dataset~\cite{lu2016visual}. They also provide a dataset which led to extensive studies on visual relationship. There are two directions in the study of VRD: one way is to solve the long-tail distribution of the label by leveraging semantic information such as language prior~\cite{plummer2016phrase,yu2017visual}, and the other way is to improve the representation power of the model by devising better architecture such as the attention model~\cite{dai2017detecting,li2017vip,yang2018shuffle,yin2018zoom,zhang2017visual,zhang2017relationship,zhuang2017towards}. Previous HOI detection works mostly focused only on the second direction~\cite{}. Recently, there has been works to attempt to exploit language prior to solve HOI detection~\cite{}, and we also focus on this direction.

\begin{figure*}[t]
    \centering
    \includegraphics[width=0.32\linewidth,keepaspectratio]{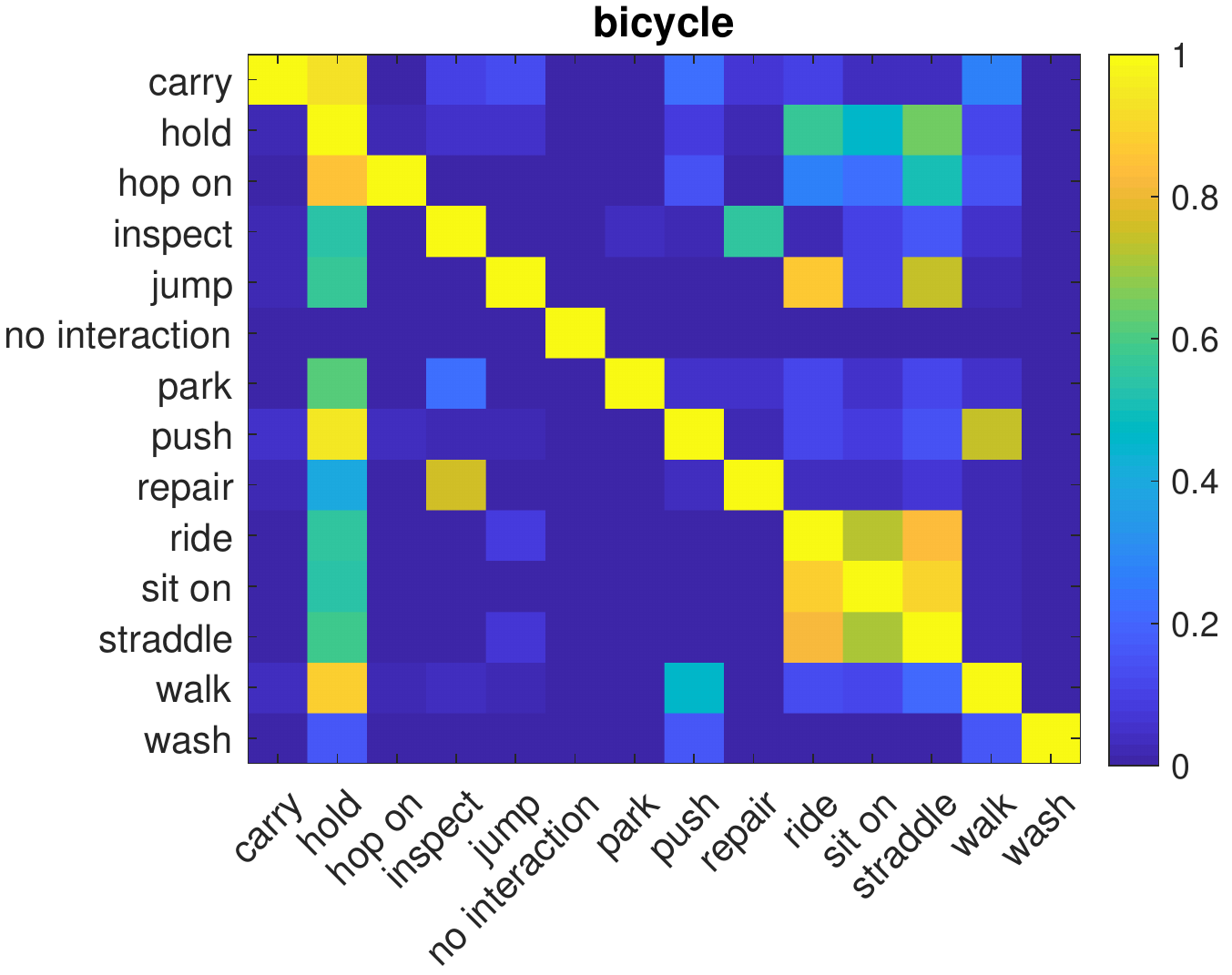}
    \includegraphics[width=0.32\linewidth,keepaspectratio]{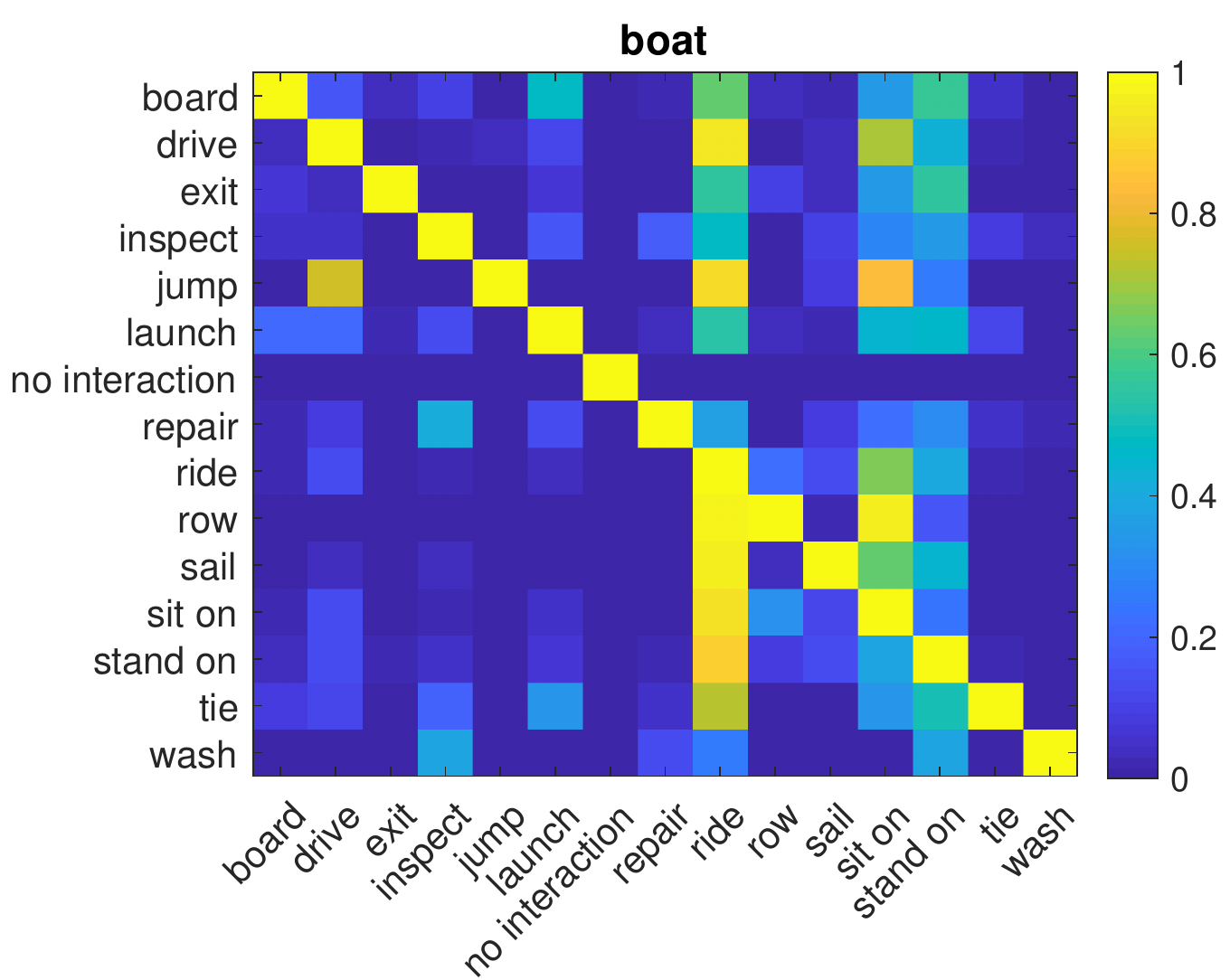}
    \includegraphics[width=0.32\linewidth,keepaspectratio]{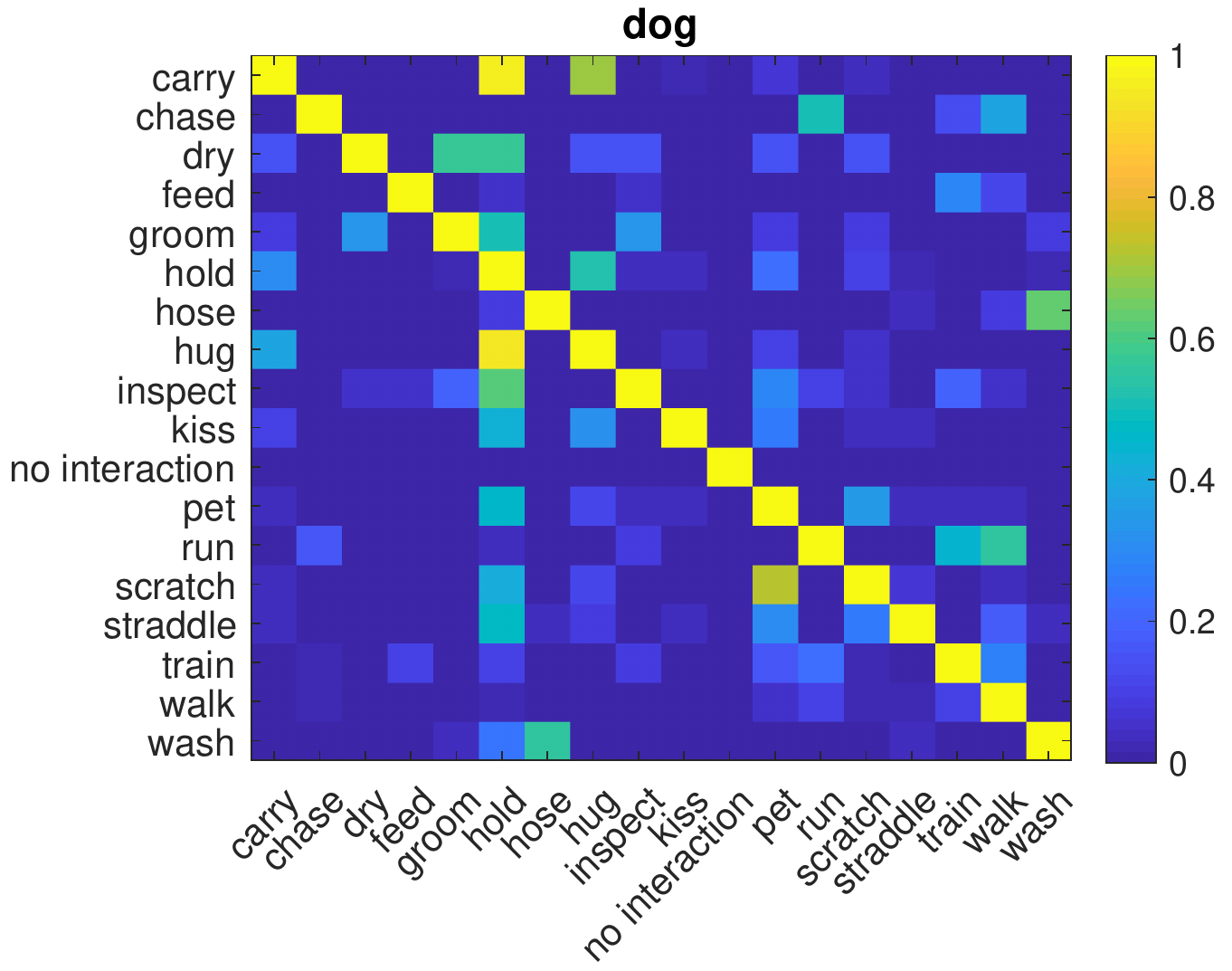}
    \vspace{-2mm}
    \caption{Examples of co-occurrence matrices constructed for several objects (bicycle, boat, dog). %constructed from HICO-Det dataset. 
    Along the Y-axis is the given action, and the X-axis enumerates conditional actions. 
    Each element represents the conditional probability that an action occurs when another action is happening.
    %The conditional probabilities can be categorized into three different types: prerequisite, exclusion, and overlapping.
    }
    \vspace{-2mm}
    \label{fig:co-occurrence}
\end{figure*}

\noindent\textbf{Label Hierarchy in Multi-label Learning} 
The hierarchical structure of label categories has long been exploited for multi-label learning in various vision tasks, \eg~image/object classification~\cite{deng2014large,yan2015hd}, detection~\cite{hwang2011sharing,marszalek2007semantic}, and human pose estimation~\cite{sun2019explicit,sun2015cascaded}. In contrast, label hierarchy has rarely been considered in HOI detection. Inspired by previous work~\cite{deng2014large} that uses Hierarchy and Exclusion (HEX) graphs to encode flexible relations between object labels, we introduce the first method to take advantage of an action label hierarchy for HOI recognition.
{While label hierarchies have commonly been used, our method is different in that it is defined by co-occurrences (rather than privileged semantics or a taxonomy~\cite{deng2014large}). This co-occurrence based hierarchy can be determined statistically, without direct human supervision.}

\section{Proposed Method}

{Our method for utilizing the co-occurrence information of HOI labels consists of three key components:} 
(1) establishing action co-occurrence priors (\Sref{sec.prior}), (2) hierarchical learning including anchor action selection (\Sref{sec.anchor}) and devising the hierarchical architecture (\Sref{sec.architecture}), followed by an extension of the architecture via self-attention (\Sref{sec.attention}), and 
(3) {ACP projection for knowledge distillation} (\Sref{sec.knowledge_distillation}), followed by an extension of the loss function via a language prior (\Sref{sec.language}).

\subsection{Action Co-occurrence Priors}
\label{sec.prior}

{Here, we formalize the action co-occurrence priors.} 
The priors for the actions are modeled by a co-occurrence matrix $C \in \mathbb{R}^{N\times N}$ where an entry $c_{ij}$ in $C$ represents the conditional probability that action $j$ occurs when action $i$ is happening: 
\begin{equation}
    c_{ij} = p(j|i), i,j \in [0, N),
\label{eq:acp}
\end{equation}
where $N$ 
(117 for the HICO-Det~\cite{chao2018learning} dataset) 
denotes the total number of action classes and $i,j$ are indices of two actions. 
By definition, the diagonal entries are one.
$C$ is constructed from the target HOI detection dataset by counting the {image-level} statistics of its training labels.
%\djkim{Note when constructing co-occurrence matrix, we don't need any data from external sources.} \steve{Suggest removing this sentence}
Examples of co-occurrence matrices constructed for a single object are visualized in \Fref{fig:co-occurrence}.

Meanwhile, we also consider the complementary event of action $i$ (\ie~where the $i$-th action does not occur) and denote it as $i^{'}$, such that $p(i^{'}) + p(i) = 1$.
The complementary action co-occurrence matrix $C^{'} \in \mathbb{R}^{N\times N}$ can thus be defined by entries $c_{ij}^{'}$ in $C^{'}$ that represent the conditional probability that an action $j$ occurs when another action $i$ does not occur: 
\begin{equation}
    c^{'}_{ij} = p(j|i^{'}), i,j \in [0, N).
\end{equation}
In this matrix, the diagonal entries are zero.

It can be seen from \Fref{fig:co-occurrence} that different types of relationships can exist between actions. 
They can be divided into three forms.
The first is the \textbf{prerequisite} relationship, where the given action is highly likely to co-occur with the conditional action. 
For example, the HOI `sit on-bicycle' is a prerequisite of the HOI `ride-bicycle'. 
In this case, 
%the action co-occurrence prior is strong, such that 
$p(\text{sit on-bicycle}|\text{ride-bicycle})$ is close to $1$.
Next is \textbf{exclusion}, where the given action is highly unlikely to co-occur with the conditional action. An example is that the HOI `wash-bicycle' and HOI `ride-bicycle' are unlikely to happen together. 
As a result, $p(\text{wash-bicycle}|\text{ride bicycle})$ is close to $0$.
Finally, we have \textbf{overlapping}, where the given action and conditional action may possibly co-occur, for example HOI `hold-bicycle' and HOI `inspect-bicycle', such that $p(\text{hold-bicycle}|\text{inspect-bicycle})$ is in between $0$ and $1$.
We introduce two 
%different 
ways to exploit the co-occurrence matrix by regarding exclusion (action space partitioning) and prerequisite actions (knowledge distillation). 
Detailed descriptions of the proposed methods are presented in the following sections.

The strong relationships that may exist between action labels can provide strong priors on the presence or absence of an HOI in an image. 
{In contrast to previous works where models may implicitly learn label co-occurrence via relational architectures~\cite{baradel2018object,zhang2019co}, we explicitly exploit these relationships between action labels as priors, to effectively train our model especially for rare HOIs.}

%%%%%%%%%%%%%%%%%%%%%%%%%%%%%%%%%%%%%%%%%%%%%%%%%%%%%%%%%%%%%%%%%%%%%%%%%%%%%%%%%%%%%%%%%%%%%%%%%%%%%%%%%%%%%%%%%%
%\subsection{Hierarchical Learning with Action Space Partition}
%\label{sec.hierarchical_learning}

\subsection{Anchor Action Selection via Non-Exclusive Suppression}
\label{sec.anchor}
% "HD-CNN: Hierarchical Deep Convolutional Neural Networks for Large Scale Visual Recognition" ("In image classification, visual separability between different object categories is highly uneven, and some categories are more difficult to distinguish than others. ")
{{From a co-occurrence matrix for an object, one can see that some actions are close in semantics or commonly co-occur while others are not. }}
Intuitively, 
%closely related actions (\eg~`sit on-bicycle' and `straddle-bicycle') 
visually similar actions (\eg~`ride-horse' and `hop on-horse')
tend to be harder to distinguish from each other, but these actions should not occur at the same time.
{If the positive labels for these actions are rare, then they become even more difficult to distinguish.}
Such cases require fine-grained recognition~\cite{fu2017look} and demand more dedicated classifiers.
This motivates us to learn HOI classes in a coarse-to-fine manner. 
In particular, as a pre-processing step, we first collect a set of mutually exclusive action {classes}, called \emph{anchor actions}, which tend to be distinguishable from one another. 
The anchor actions are used to partition the entire action {label} space into fine-grained sub-spaces. The other action {classes} are attributed to one or more sub-spaces and recognized in the context of a specific anchor action.
{In summary, unlike previous HOI detection works which predict action probabilities independently of one another, we divide the whole action label set into two sets, one for anchor actions and one for regular actions, which are modeled in different ways as explained in detail in \Sref{sec.architecture}}.

% \djkim{
% As the first way to exploit $C$, we cluster the actions into several groups of actions which are exclusive to each other. 
% In particular, if $c_{ij}=0$, the action $i$ and $j$ must belong to different groups. We check the correlation of the verbs one by one, and we distribute the verb classes into $G$ number of disjoint action sets. 
% The detailed process is as follows: 
% (1) First find verbs with no co-occurrence with other verbs. These will be the groups with only one action. \js{($\Sigma_j c_{ij}=1$ (?))}
% (2) Next, find verbs with a co-occurrence with only one verb. These verbs are the furthest verbs with other verbs in the co-occurrence graph. Thus these verbs will be the anchors of each groups.\js{($\Sigma_j c_{ij}=2$ (?))}
% (3) Next, assign the remaining verbs to the groups. If a verb has co-occurrence with a group the verb will be assigned to the group.\js{($\Sigma_j c_{ij}>1$ (?))}
% (3-1) If a verb has co-occurrence with multiple groups, assign the verb is assigned to the group with higher co-occurrence probability.
% }

Before 
%we train a model
training, in selecting anchor actions, we seek a set of action {classes} that are exclusive of one another. To this end, we define the exclusiveness of an action {class} as counting the number of actions that never occur if action $i$ is happening: 
\begin{equation}
e_i = \sum_j ({1} \text{ if } (c_{ij} =0), \text{ else } {0}).
\label{eq:exclusiveness}
\end{equation}
$e_i$ will have a high value if few other actions can occur when $i$ does.
Based on exclusiveness values, the anchor action {label} set $\mathcal{D}$ is generated through \emph{\textbf{non-exclusive suppression (NES)}} as described in \Aref{alg:correlation}. It iteratively finds the most exclusive action {class} as an anchor action and removes remaining action {classes} that are not exclusive to the selected anchor actions. 
%\Tref{table:anchor} lists the maximum number of anchor actions from the HICO-Det dataset in the order in which they are selected.
%For example, when setting the number of anchor actions to be 10, we take from the first action (`flush') to the tenth action (`teach') as the set of anchors, and the rest are associated to `other.'
The anchors in the list are action {classes} that never occur together in the training labels. For example, if an action such as `toast' is on the anchor list, then actions like `stand' and `sit' cannot be on the list because they may co-occur with `toast', while actions such as `hunt' or `hop on' can potentially be on the list.
While there may exist other ways the anchor action selection could be done, we empirically found this approach to be simple, effective in terms of detection accuracy, and computationally efficient (less than 0.01 second).

\begin{algorithm}[ht]
\KwIn{
$\mathcal{E}=\{e_{i}, i \in [0, N)\}, \mathcal{C}=\{c_{ij}, i,j \in [0, N)\}$\;
}
\KwOut{
$\mathcal{D}$
} 
\Begin
{
  $\mathcal{D} \gets \{\}$  \;
  \While{$\mathcal{E}$ is not empty}
%   \If{$len(\mathcal{D}) < G$}
  {
    \# Find the most exclusive action\;
    $m \gets argmax \mathcal{E}$\;
    %\# \xiao{This is always true.}\;
    %\If{$\sum_{i=1}^{length(D)} c_{m\mathcal{D}_i}==0$}
   % {
       % \# If exclusive, then add to the anchor list\;\djkim{(condition added)}
   $\mathcal{D} \gets \mathcal{D} \cup \{m\}$  \;
   % }
    \For{$e_k \in \mathcal{E}$}{
            \# Remove the actions correlated (not exclusive) to $m$\;
            \If{$c_{mk} > 0$}
            {
                $\mathcal{E} \gets \mathcal{E} - e_k; \mathcal{C} \gets C-\{c_{ij}, i \text{ or } j = k\}$
            }
    }
  }
}
\caption{
{Non-Exclusive Suppression (NES) algorithm for mutually exclusive anchor action selection.
}
}
\label{alg:correlation}
\end{algorithm}

The anchor action {label} set acts as a finite partition of the action {label} space (a set of pairwise disjoint events whose union is the entire action {label} space). To form a complete action {label} space, we add an \emph{\textbf{`other'}} anchor action, denoted as $\mathcal{O}$, for when an action {class} does not belong to $\mathcal{D}$. 
Finally, we have $|\mathcal{D}| + 1$ anchor actions including $\mathcal{D}$ and the `other' action {class} $\mathcal{O}$. 
% (\ie $|\mathcal{D}\cup\mathcal{O}|=d$)

There are several benefits to having this anchor action {label} set.
First, only one anchor action can happen at one time between a given human and object. Thus, we can use the relative (one-hot) probability representation with \emph{softmax} activation instead of a sigmoid, where softmax features were shown to compare well against distance metric learning-based features~\cite{horiguchi2019significance}.
Second, anchor actions tend to be easier to distinguish from one another since they generally have prominent differences in an image.
Third, it decomposes the action {label} space into several sub-spaces, which facilitates a coarse-to-fine solution. Each sub-task will have a much smaller solution space, which can improve learning.
Finally, each sub-task will use a standalone sub-network which focuses on image features specific to the sub-task, which is an effective strategy for fine-grained recognition~\cite{fu2017look}.

After selecting anchor actions, the entire action {label} set $\mathcal{A}$ is divided into the anchor action {label} set $\mathcal{D}$ and the remaining set of `regular' action {classes} $\mathcal{R}$, so that $\mathcal{A} = \{\mathcal{D}, \mathcal{R}\}$.
Each of the regular action {classes} is then associated with one or more anchor actions to form $|\mathcal{D}|+1$ action groups $\mathbf{G} = \{\mathcal{G}_i; i \in \mathcal{D}\cup\mathcal{O}\}$, one for each anchor action. 
% \begin{equation}
%     \mathbf{G} = \{\mathcal{G}_i; i = 1,...,d\}.
% \label{eq:action_groups}
% \end{equation}
A regular action {class} $j \in \mathcal{R}$ will be assigned to the group associated with anchor action $i$ ($\mathcal{G}_i$) if action $j$ is able to co-occur with the anchor action $i$,
\begin{equation}
    j \in \mathcal{G}_i,  \text{if } c_{ij} > 0 \;\; (i \in \mathcal{D}\cup\mathcal{O}, j \in \mathcal{R}).
\label{eq:action_groups2}
\end{equation}
Note that the anchor actions themselves are not included in the action groups and a regular action $j$ can be assigned to multiple action groups since it may co-occur with multiple anchor actions.

%\begin{table}[t]
% \centering
% \caption{List of 54 anchors extracted from among the 117 actions of the HICO-Det dataset.}
% \resizebox{1\linewidth}{!}{%
% \begin{tabular}{|cccccc|}
%\hline
%flush	& tag	& stab& wave	& brush with	& hunt\\
%toast	& pay	& move	& teach	& no interaction	& eat at\\
%milk	& squeeze	& greet& stop at	& stir	& install	\\
%point	& sign	& paint	& shear	& release	& control	\\
%break	& light	& lose	& zip	& lift	& pack	\\
%cut with	& type on	& talk on	& set	& slide	& operate	\\
%spin	& assemble	& pour	& lie on	& turn	& chase	\\
%herd	& flip	& buy	& hose	& kick	& row	\\ 
%peel	& dry	& hop on	& direct	& adjust	& kiss	\\
%\hline
%  \end{tabular}
% }
%\label{table:anchor}
% \end{table}

\begin{figure*}[t]
\centering
    \includegraphics[width=1\linewidth,keepaspectratio]{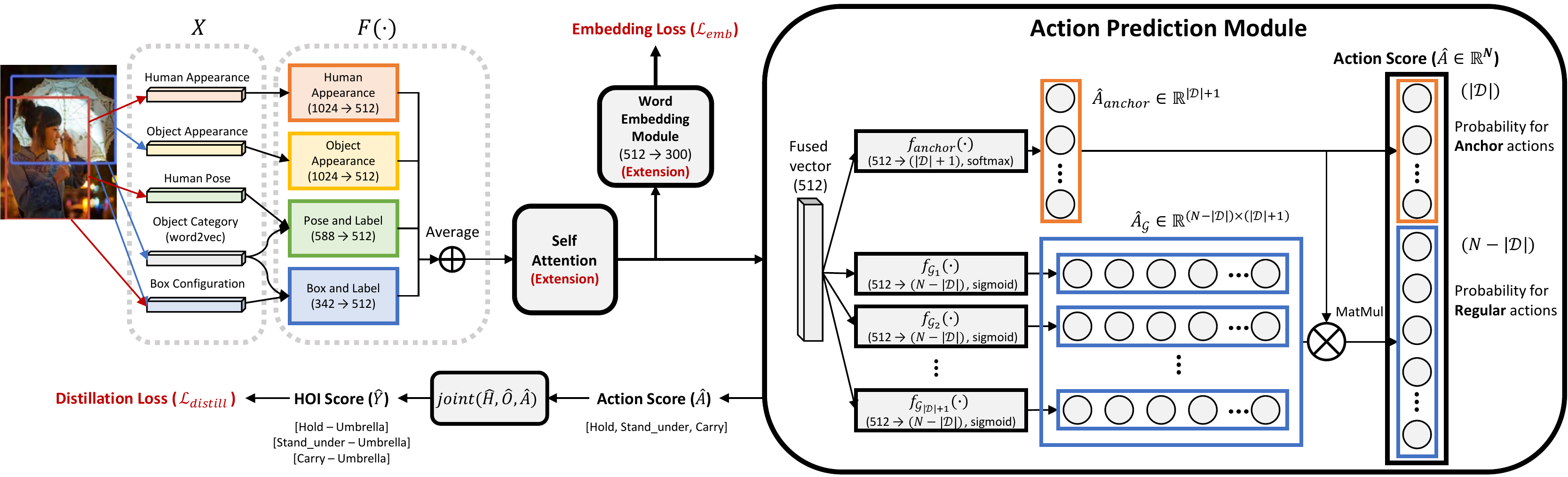}
    \vspace{-2mm}
    \caption{Illustration of our overall network architecture.
	Our work differs from the baseline \cite{gupta2019no} by the addition of a hierarchical \emph{action prediction module}.
	For our hierarchical architecture, anchor action probabilities are directly generated by a \emph{softmax} sub-network. Regular action probabilities are generated by a matrix multiplication of the anchor probability and the output from a few \emph{sigmoid} based conditional sub-networks.}
	\vspace{-2mm}
    \label{fig:model_merged}
\end{figure*}

\subsection{Hierarchical Architecture}
\label{sec.architecture}
We implement our hierarchical approach upon several existing HOI detection architectures~\cite{gao2020drg,gupta2019no,liao2020ppdm}.
In this section, we introduce our main baseline architecture, built on the `No-Frills' (NFs) baseline presented in~\cite{gupta2019no} on account of its simplicity, effectiveness, and code availability~\cite{nofrillsgit}.
%We implemented our hierarchical approach upon the `No-Frills' (NFs) baseline presented in~\cite{gupta2019no} on account of its simplicity, effectiveness, and code availability~\cite{nofrillsgit}. 
Here, we give a brief review of the NFs architecture. 

\noindent\textbf{Baseline Network}
%\djkim{(R1 tackled that this paragraph is repetitive and long-winded.)} \steve{Revised.}
NFs follows the common scheme of CNN feature extraction followed by multi-information fusion. The feature extraction uses an off-the-shelf Faster R-CNN~\cite{ren2015faster} object detector with ResNet152~\cite{he2016deep} backbone network to detect human and object bounding boxes. From the final fully connected (FC) layer of the detector, the features $\hat{x}=\{\hat{x}_h,\hat{x}_o\}$ are extracted, where $\hat{x}_h$ and $\hat{x}_o$ denote human and object appearance, respectively. These CNN features, together with the human pose ($\hat{k}$), object category ($\hat{o}$) and bounding boxes ($\hat{b}$ including both object and human bounding boxes) compose the multiple information as $X=\{\hat{x},\hat{k},\hat{o},\hat{b}\}$. As illustrated in \Fref{fig:model_merged}, they are fed to our target model via corresponding network streams: human appearance ($f_h$), object appearance($f_o$), bounding box and object category ($f_b$), and human pose and object category ($f_k$). Each individual type of information is first fed through a separate network of two FC layers to generate a fixed dimension (number of actions $N$) feature. Then, all the features are added together and sent through a \emph{sigmoid} activation to obtain the probability prediction for the action $a$:
\begin{equation}
    \hat{A}= sigmoid(F(X)) \in \mathbb{R}^N,
\label{eq:base_action_pred}
\end{equation}
where $\hat{A}(a) = p(a|X)$ represents the probability prediction for action class $a$.
The multi-information fusion procedure $F(X)$ is expressed as 
 \begin{equation}
     F(X) = f_h(\hat{x}_h) + f_o(\hat{x}_o) + f_k(\hat{k}||\hat{o})+ f_b(\hat{b}||\hat{o}),
 \label{eq:base_action_pred_simple}
 \end{equation}
 where 
 %the operation 
 $||$ denotes concatenation. The baselines built on other models~\cite{gao2020drg,liao2020ppdm} are constructed in a similar manner.

To eliminate training-inference mismatch, NFs directly optimizes the HOI class probabilities instead of separating the detection and interaction losses as done in~\cite{gao2018ican,gkioxari2018detecting}. The final HOI prediction is a joint probability distribution over $M$ number of HOI classes (600 for HICO-Det dataset)  {computed from the probabilities $\hat{H}$, $\hat{O}$, and $\hat{A}$ for human, object, and action, respectively:}
\begin{equation}
    \hat{Y} = joint(\hat{H}, \hat{O}, \hat{A}) \in \mathbb{R}^{M}.
\end{equation}
Specifically, for an HOI class $(h,o,a)$,
\begin{equation}
\begin{split}
    %\hat{Y} (h,o,a) & = \hat{H}(h) * \hat{O}(o) * \hat{A}(a) \\
    %& = p(h|I) * p(o|I) * p(a|X),\\
    \hat{Y} (h,o,a) & = \hat{H}(h) * \hat{O}(o) * \hat{A}(a) = p(h|I) * p(o|I) * p(a|X),\\
\end{split}
    \label{eq:base_hoi_pred}
\end{equation}
% \xiao{This one better?}
% \begin{equation}
%     %\hat{Y} 
%   %= joint(\hat{H}, \hat{O}, \hat{A}))\\
%   % = joint({H}, {O}, {A}) 
%     %= joint(h,o,a) 
%     \hat{Y} (h,o,a)= p(h|I) * p(o|I) * p(a|X),
%     \label{eq:base_hoi_pred}
% \end{equation}
where $\hat{H}(h)=p(h|I)$ and $\hat{O}(o)=p(o|I)$ are the probability of a candidate box pair being a human $h$ and object $o$, provided by the object detector~\cite{ren2015faster}. Finally, the binary cross-entropy loss $\mathcal{L}(\hat{Y}, Y^{gt})$ is directly computed from the HOI prediction.
%\begin{equation}
%l(\hat{Y}, Y^{gt}) = -(Y^{gt}log(\hat{Y})+(1-Y^{gt})log(1-\hat{Y})).
%\label{eq:base_loss}
%\end{equation}
This ‘No-Frills’ baseline network is referred to as \emph{\textbf{Baseline}}.

\noindent\textbf{Modified Baseline Network} 
{For a stronger baseline comparison,}
we make two {simple} but very effective modifications on the baseline network.
(1) Replace the one-hot representation with the Glove word2vec~\cite{pennington2014glove} representation for the object category ($\hat{o}$ ).
(2) Instead of directly adding up the multiple information, we average them and forward this through another \emph{action prediction module}
%  $f_{sub}$  of a few FC layers 
to obtain the final action probability prediction. As a naive implementation of the \emph{\textbf{Modified  Baseline}}, we simply use a sub-network $f_{sub}$ of a few FC layers as the \emph{action prediction module}.
Then \Eref{eq:base_action_pred}-\ref{eq:base_action_pred_simple} are modified to
\begin{equation}
\hat{A} = 
sigmoid(f_{sub}(F(X))),
 \label{eq:modified}
\end{equation}
 \begin{equation}
 F(X) = \dfrac{f_h(\hat{x_h}) + f_o(\hat{x_o}) + f_k(\hat{k}||\hat{o})+ f_b(\hat{b}||\hat{o})}{n_{stream}}.
 \end{equation}
 In this case, $n_{stream}=4$. 

Our hierarchical architecture further modifies the \emph{action prediction module} by explicitly exploiting ACP information as described in the next paragraph.

%\begin{figure}[t]
%	\centering
%	\subfigure[\emph{\textbf{Modified Baseline}}]{\includegraphics[width=.9\linewidth,keepaspectratio]{Figures/net_design1.pdf}}\\
%	\subfigure[\emph{\textbf{MultiTask}}]{\includegraphics[width=.9\linewidth,keepaspectratio]{Figures/net_design2.pdf}}\\
%	\subfigure[\emph{\textbf{TwoStream}}]{\includegraphics[width=.9\linewidth,keepaspectratio]{Figures/net_design22.pdf}}\\
 %   \subfigure[\emph{\textbf{Hierarchical}}]{\includegraphics[width=.9\linewidth,keepaspectratio]{Figures/net_design3.pdf}}
%	\caption{
 %	Design choices for the action prediction module of our network. (a) \emph{\textbf{Modified Baseline}} architecture without leveraging the prior of anchor actions. (b) \emph{\textbf{MultiTask}} which utilizes the anchor actions as in a multi-task learning manner. (c) \emph{\textbf{TwoStream}} which separately predicts the anchor and the regular actions but without using the hierarchical modeling between anchor and regular actions. (d) The proposed hierarchical target (\emph{\textbf{Hierarchical}}). Anchor probability is directly used as a final score, and we exploit multiple conditional sub-networks to further compute the probabilities for the regular actions.
%    }
%	\label{fig:net_design_choice2}
%\end{figure}

\noindent\textbf{Proposed Hierarchical Architecture}
%\djkim{(Maybe we should make our methodology section easier to read, if possible, perhaps by reducing equations or notations somehow.)} \xiao{Yes, you can try to simplify this first and we can discuss if it will lose veracity. I will also think about this but I don't have a clear idea now.}
Now we introduce the \emph{action prediction module} for our hierarchical architecture (illustrated in \Fref{fig:model_merged}) that better exploits the inherent co-occurrence among actions. While the baseline network predicts all the action probabilities directly from $F(\cdot)$ with a single feed-forward sub-network $f_{sub}$, we instead use $|\mathcal{D}|+2$ sub-networks where one ($f_{anchor}(\cdot)$) is first applied to predict the anchor action set and then one of the $|\mathcal{D}|+1$ other sub-networks ($f_{\mathcal{G}_i}(\cdot)$) which corresponds to the predicted anchor action is used to estimate the specific action within the action group. Because of the mutually exclusive property of anchor actions, we use the \emph{softmax} activation for anchor action predictions, while employing the \emph{sigmoid} activation for regular action prediction conditioned on the action group:
% \begin{equation}
%     \hat{A}_{anchor}(i) = softmax(f_{anchor}(F(X))), \text{where } i \in \mathcal{D}
% \label{eq:hie_anchor_pred}
% \end{equation}
% \begin{equation}
%     \hat{A}_{\mathcal{G}_i}(j) = sigmoid(f_{\mathcal{G}_i}(F(X))), \text{where } i \in \mathcal{D}, j \in \mathcal{G}_i,
% \label{eq:hie_group_pred}
% \end{equation}

\begin{equation}
    \hat{A}_{anchor} = softmax(f_{anchor}(F(X))) \in \mathbb{R}^{|\mathcal{D}|+1}
\label{eq:hie_anchor_pred}
\end{equation}
\begin{equation}
    \hat{A}_{\mathcal{G}_i} = sigmoid(f_{\mathcal{G}_i}(F(X))) \in \mathbb{R}^{N-|\mathcal{D}|}, \text{where } i \in \mathcal{D}\cup\mathcal{O},%i \in \mathcal{D}\cup \mathcal{O},
\label{eq:hie_group_pred}
\end{equation}
where $\hat{A}_{anchor}(i)$ is directly used as the final probability predictions for the anchor actions:% ($p(i|X) = \hat{A}_{anchor}(i), i \in \mathcal{D}$).
\begin{equation}
     p(i|X) = \hat{A}_{anchor}(i), i \in \mathcal{D}.
\end{equation}
We let $\hat{A}_{\mathcal{G}_i}(j)$ represent the learned conditional probability that action $j$ occurs when action $i$ is happening, namely, 
%($p(j|i,X) = \hat{A}_{\mathcal{G}_i}(j)$).
\begin{equation}
 \hat{A}_{\mathcal{G}_i}(j) = p(j|i,X).
% p(j|i,X) = \hat{A}_{\mathcal{G}_i}(j).
\end{equation}
Since the anchor action set is a finite partition of the entire action {label} space, the probability of a regular action $j$ is predicted according to the law of total probability:
\begin{equation}
    \begin{split}
    \hat{A}_{regular}(j) = p(j|X) & 
    =\sum_{ i \in \mathcal{D}\cup\mathcal{O}}^{} p(i|X) * p(j|i,X) \\
         & = \sum_{ i \in \mathcal{D}\cup\mathcal{O}}^{} \hat{A}_{anchor}(i) * \hat{A}_{\mathcal{G}_i}(j),% \text{where } j \in R.%, i \in \mathcal{D}.
    \end{split}
\label{eq:hie_regular_pred}
\end{equation}
where $j \in \mathcal{R}$.
% For an anchor action (\eg $a \in \mathcal{D}$), $ p(a|i,X)=1$ if $i=a$ and $ p(a|i,X)=0$ otherwise.
Thus, 
instead of \Eref{eq:modified},
%combining Equations~\ref{eq:hie_anchor_pred}, ~\ref{eq:hie_group_pred} and ~\ref{eq:hie_regular_pred}, 
we obtain the final action probability predictions for our hierarchical architecture $\hat{A}(a) = p(a|X)$ as 
\begin{equation}
    \hat{A}(a) = %p(a|X)= 
\begin{cases}
    \hat{A}_{anchor}(a),
    %softmax(f_{anchor}(F(X))),
    & \text{if } a\in \mathcal{D}\\
    \sum_{ i \in \mathcal{D}\cup\mathcal{O}}^{} \hat{A}_{anchor}(i) * \hat{A}_{\mathcal{G}_i}(a),
    %sigmoid(f_{\mathcal{G}_i}(F(X))),              
    & \text{otherwise.}
\end{cases}
\label{eq:hie_action_pred}
\end{equation}
We use the same method as in \Eref{eq:base_hoi_pred} and a cross-entropy loss $\mathcal{L}$ to compute the final HOI probability prediction $\hat{Y}$.

To demonstrate the effectiveness of the hierarchical learning, we introduce two other baselines,  \emph{\textbf{MultiTask}} and \emph{\textbf{TwoStream}}, that lie between the \emph{\textbf{Modified Baseline}} and our hierarchical learning. \emph{\textbf{MultiTask}} only uses the anchor action classification as an additional multi-task element to the \emph{\textbf{Modified Baseline}}. %
\emph{\textbf{TwoStream}} separately predicts the anchor and the regular actions
but without using the hierarchical modeling between anchor and regular actions. 
%We show an illustration of these methods in \Fref{fig:net_design_choice2}.

\subsection{Self-Attention Module Extension}
\label{sec.attention}
%\djkim{(extension contents about architecture)}
%Since our network only utilizes the features from a given human-object pair, it may lack a global understanding of the {constituent} objects {in the entire image}, \ie~global context.
In order to enhance the capability for holistic relational understanding across objects or humans, we additionally leverage a self-attention module that applies the non-local layer~\cite{wang2018non} to each human-object pair. %, different from the approach of Wang~\etal~\cite{wang2018non} that applies it densely to the feature map.
%This self-attention module enhances the relational information across all human-object pairs via the attention mechanism.
In particular, let $Z \in \R^{B\times 512}$ denote a stack of $B$ merged human-object features from our $F(\cdot)$ ($Z=F(X)$).
Then, we compute the relational association matrix by:
\begin{equation}
        R = \mathrm{Softmax}({\sigma}(Z W_a){\sigma}(Z W_b)^\top) \in \R^{B\times B},
		\label{eq:REM1}
\end{equation}
where {$\sigma(\cdot)$ denotes ReLU and} $W_a, W_b \in \R^{512\times 128}$ are learnable {weights} that map the human-object features $Z$ to their own roles, (\eg~key and query) and 
the softmax operation is applied row-wise.
Then, the relational feature matrix is computed by:
\begin{equation}
A = R\sigma(ZW_x)W_z^\top \in \R^{B\times 512},
\label{eq:REM2}
\end{equation}
where $W_x\in \R^{512\times 128}$ and $W_z\in \R^{512\times 128}$ are again learnable {weights}.
The matrix $A$ encodes aggregated features across all the objects according to the degree of relational association given in $R$. %, which is similar to message passing that exchanges information according to the relationship.
This relational feature matrix is combined with the original feature $Z$ by $\tilde{Z} = Z + A$, so that $Z$ is enhanced by holistic relational information.
%This can be viewed as augmenting it with richer semantic meaning, \eg~`petting-bird' ($Z$) may be augmented from a bird that another person is watching or a bird that is fed by a person, depending on the surroundings.
%Also, it is akin to the residual connection, allowing efficient training via the residual learning mechanism~\cite{he2016deep}.
%Different from the non-local approaches \cite{wang2018non,woo2018linknet}, we introduce non-linear activations, ReLU, in Eqs.~(\ref{eq:REM1}) and (\ref{eq:REM2}), motivated by a low-rank bilinear pooling~\cite{kim2016hadamard}.
%We empirically found that this modification leads to noticeable performance improvement.

%We introduce a hierarchical network architecture to better exploit the inherent co-occurrence property between actions (illustrated in \Fref{fig:architecture}). 
%We first use a sub-network $\mathcal{N}_{softmax}$ that takes the fusion feature $f$ as input and outputs the probabilities of the anchor actions in $D$. 
%Since the anchor actions are exclusive to each other, we use the \emph{softmax} activation for the final FC layer instead of \emph{sigmoid}.

%Then, we add $G$ sub-networks $\{N_{sigmod}^i, i \in G\}$ to $f$ for different action groups. 
%Since the actions within a group have high probabilities of co-occurrence, we use the \emph{sigmod} activation for the final FC layer of $N_{sigmod}^i$.

%%%%%%%%%%%%%%%%%%%%%%%%%%%%%%%%%%%%%%%%%%%%%%%%%%%%%%%%%%%%%%%%%%%%%%%%%%%%%%%%%%%%%%%%%%%%%%%%%%%%%%%%%%%%%%%%%%%
\subsection{{ACP Projection for Knowledge Distillation}}
\label{sec.knowledge_distillation}

Knowledge distillation~\cite{hinton2015distilling} was originally proposed to transfer knowledge from a large network to a smaller one. 
Recently, knowledge distillation has been utilized for various purposes such as lifelong learning~\cite{li2016learning}, multi-task learning~\cite{kim2018disjoint} or modality transfer~\cite{cho2021dealing,hoffman2016learning}.
Hu~\etal~\cite{hu2016harnessing} extended this concept to distill prior knowledge in the form of logic rules into a deep neural network. Specifically, they propose a teacher-student framework to project the network prediction (student) to a rule-regularized subspace (teacher), where the process is termed distillation. 
The network is then updated to balance between emulating the teacher's output and predicting the true labels.

Our work fits this setting as
%because 
the ACPs can act as a prior to distill. 
We first introduce \emph{ACP Projection} to map the action distribution to the ACP constraints. Then, we use the teacher-student framework~\cite{hu2016harnessing} to distill knowledge from ACPs.

\noindent\textbf{ACP Projection} In ACP Projection, an arbitrary action distribution 
$A = \{p(i), i \in [0, N)\} \in \mathbb{R}^{ N}$
is projected into the ACP-constrained probability space:
\begin{equation}
A^{*} = project(A, C, C^{'}) \in \mathbb{R}^{N},
\end{equation}
where $A^{*}$ is the projected action prediction. The projected probability for the $j$-th action $A^{*}(j) = p(j^{*})$ is generated using the law of total probability:
\begin{equation}
    \begin{split}
    p(j^{*}) & =\dfrac{1}{N}\sum_{i=1}^{N} (p(i) * p(j|i) + p(i^{'}) * p(j|i^{'})) \\
        %  & = \dfrac{1}{N} \sum_{i=1}^{N} (p(i) * c_{ij} + p(i^{'}) * c_{ij}^{'}) \\
         & = \dfrac{1}{N} (\sum_{i=1}^{N} p(i) * c_{ij} + \sum_{i=1}^{N} (1-p(i)) * c_{ij}^{'}). \\
    \end{split}
\end{equation}

In matrix form, the ACP projection is expressed as
\begin{equation}
project(A, C, C^{'}) = \dfrac{AC + (1-A)C^{'}}{N}.
\end{equation}

In practice, we use the object-based action co-occurrence matrices $C_{o}\in\mathbb{R}^{N\times N}$ and $C_{o}^{'}\in\mathbb{R}^{N\times N}$ which only count actions related to a specific object $o$. \Fref{fig:co-occurrence} shows examples of $C_{o}$ with respect to object classes.
Also, we give different weights $\alpha$ and $\beta$ {as hyper-parameters} to the action co-occurrence matrix $C_{o}$ and its complementary matrix $C_{o}^{'}$, with the weights subject to $\alpha + \beta = 2,\alpha > \beta$.
The projection function is then modified to
\begin{equation}
project(A, C_{o}, C^{'}_{o}) = \dfrac{\alpha AC_{o} + \beta (1-A)C^{'}_{o}}{N}.
\end{equation}
This is done because we empirically found the co-occurrence relationships in $C_{o}$ to generally be much stronger than the complementary actions in $C^{'}_{o}$.

\noindent\textbf{Teacher-Student Framework} Now we can distill knowledge from the ACPs using ACP Projection in both the training and inference phases. There are three ways ACP Projection can be used:
(1) Directly project the action prediction $\hat{A}$ into the ACP-constrained probability space at the testing phase to obtain the final action output (denoted as \emph{\textbf{PostProcess}}). 
(2) Project the action prediction $\hat{A}$ in the training phase and use the projected action as an additional learning target~\cite{hu2016harnessing,yu2017visual}.
(3) Project the ground truth label $H^{gt}$, $O^{gt}$, and $A^{gt}$\footnote{The triplet ground truth labels $H^{gt}$, $O^{gt}$, and $A^{gt}$ are straightforward to determine from the HOI ground truth label $Y^{gt}$.} to the ACP space in the training phase and use the projected action $project(A^{gt}, C_{{O}^{gt}}, C_{{O}^{gt}}^{'})$ as an additional learning target.
The second and third items are incorporated into the teacher-student framework as terms in a new objective function {(denoted as \emph{\textbf{Distillation}})}:
\begin{equation}
\mathcal{L}_{distill} = \lambda_1 \mathcal{L}(\hat{Y}, Y^{gt}) + \lambda_2 \mathcal{L}(\hat{Y}, \hat{Y}_{projO}) + \lambda_3 \mathcal{L}(\hat{Y}, Y^{gt}_{projO}),
\label{eqn:distill_loss}
\end{equation}
where
\begin{equation}
\hat{Y}_{projO} = joint(\hat{H}, \hat{O}, project(\hat{A}, C_{\hat{O}}, C_{\hat{O}}^{'}))  \in \mathbb{R}^{M},
\label{eqn:teacher2_self}
\end{equation}
\begin{equation}
Y^{gt}_{projO} = joint({H}^{gt},{O}^{gt},project(A^{gt}, C_{{O}^{gt}}, C_{{O}^{gt}}^{'}))  \in \mathbb{R}^{M}.
\label{eqn:teacher2_gt}
\end{equation}
$\lambda_1, \lambda_2, \lambda_3$ are balancing weights among the ground truth HOI term and the teacher objectives.
The object type can be easily determined from the object probability predictions $\hat{O}$ or the ground truth label ${O}^{gt}$.
With the new objective function, rare classes receive different supervisions, so confusion between them can be alleviated.

\subsection{Word Embedding Loss Extension}
\label{sec.language}
%\djkim{(extension contents about loss function)}
As an extension to enhance performance, we add another loss term, namely a word embedding regression loss.
Given a word2vec representation for the object category ${\hat{o}}$, we regress our model to the same object representation in the output by computing the regressed word embedding $v$. 
Similar to Peyre~\etal~\cite{peyre2019detecting}, we design the word embedding regularization loss as the sigmoid loss function:
\begin{equation}
    \mathcal{L}_{emb}=\sum^{|O|}_{i=1}{1}_{(y^o=i)}\log(\frac{1}{1+e^{-\hat{o}\cdot v}}),
    \label{eqn:emb_loss}
\end{equation}
where $|O|$ is the number of objects (80 for HICO-Det datasets) and $y^o\in\mathbb{R}^{|O|}$ is the class for the object.
%\djkim{(We shall discuss about the notation.)}
Multi-task learning with the object prediction works as prior knowledge for action prediction, which leads to performance improvements.
Therefore, the total loss function can be described as a weighted sum of the knowledge distillation loss and the embedding loss:%function:
\begin{equation}
    \mathcal{L}_{total}=\mathcal{L}_{distill} + \lambda_0 \mathcal{L}_{emb},
    \label{eqn:total_loss}
\end{equation}
where $\lambda_0$ is the additional balancing weight between \Eref{eqn:distill_loss} and \Eref{eqn:emb_loss}.
Note that, as we have an object word2vec vector as input, this word embedding regression loss can be seen as an auto-encoding loss which is commonly used as a regularizer.

\section{Experiments}

{The goal of the experiments is to show the effectiveness and generalizability of our method.
In particular, we show that our method can consistently alleviate the long-tailed distribution problem in various setups by improving performance, especially for rare HOI classes.}
In this section, we describe the experimental setups and competing methods, and provide extensive performance evaluations and analysis of HOI detection with both quantitative and qualitative results.

\subsection{Datasets and Metrics}
\label{sec.dataset}
We evaluate the performance of our model on two popular HOI detection benchmark datasets, \textbf{HICO-Det}~\cite{chao2018learning} and \textbf{V-COCO}~\cite{gupta2015visual}.
HICO-Det~\cite{chao2018learning} extends the HICO (Humans Interacting with Common Objects) dataset~\cite{chao2015hico} which contains 600 human-object interaction categories for 80 objects. 
Different from the HICO dataset, HICO-Det additionally contains bounding box annotations for humans and objects of each HOI category. 
The vocabulary of objects matches the 80 categories of MS COCO~\cite{lin2014microsoft}, and there are 117 different verb (action) categories. 
The number of all possible triplets is $117\times80$, but the dataset contains positive examples for only 600 triplets. 
%All of these triplets are seen at least once in training.
The training set of HICO-Det contains 38,118 images and 117,871 HOI annotations for 600 HOI classes. 
The test set has 9,658 images and 33,405 HOI instances.

For evaluation, HICO-Det uses the mean average precision (mAP) metric.
Here, an HOI detection is counted as a true positive if the minimum of the human overlap IOU and object overlap
IOU with the ground truth is greater than 0.5. 
Following \cite{chao2018learning}, HOI detection performance is reported for three different HOI category sets: (1) all 600 HOI categories (Full), (2) 138 categories with fewer than 10 training samples (Rare), and (3) the remaining 462 categories with more than 10 training samples (Non-rare).
Two evaluation settings are examined: `Default' and `Known Object'. For the `Known Object' setting, we evaluate the detection only on the images containing the target object category (\eg~ ``bike'') given the HOI class (\eg~``riding-bike'').

\begin{table}[t]
\centering
    \caption{Ablation study on the HICO-Det dataset. Our final model (ACP++) that includes both hierarchical architecture and distillation followed by post processing, along with the two technical extensions of word embedding loss and self-attention, shows the best performance among the baselines.}
    \resizebox{1\linewidth}{!}{
        \begin{tabular}{l| ccc}
    		\hline
    			&		Full&Rare&Non-rare\\
    		\hline
    		\hline
    		Baseline	  & 17.56&   	13.23&   18.85 \\
    		%Modified Baseline &   19.19	& 13.40  & 21.20\\
    		Modified Baseline &   19.09	& 13.09  & 20.89\\
    		\hline
    		+Hierarchical only &   20.03	&14.52   &21.67\\
    		+Distillation only &   19.98	&13.67   &21.86\\
    		+Hierarchical+Distillation &   {20.25}	& {15.33}   &{21.72}\\
    		%\hline
    		+Hierarchical+Distillation+Post (Ours, ACP)&   \textbf{20.59}	& \textbf{15.92}   & \textbf{21.98}\\
    		\hline
    		+Hierarchical+Distillation+$\mathcal{L}_{emb}$  & 20.76&\textbf{16.35}&22.08\\
    		+Hierarchical+Distillation+SA  & 20.89&14.43&22.81\\
    		+Hierarchical+Distillation+$\mathcal{L}_{emb}$+SA (ACP++) & 20.96&15.12&22.70\\
    		+Hierarchical+Distillation+$\mathcal{L}_{emb}$+SA (ACP++) +Post &\textbf{21.27}&{15.41}&\textbf{23.02}\\
    		\hline
		\end{tabular}
	}
\label{table:ablation}
\end{table}%

\begin{table}[b]
\centering
    \caption{Performance of our models with different architectures for action prediction module. Our %related to action groups. 
    %(A):Modified baseline, (B):Multi-task learning, (C):Two-stream approach, (D):Hierarchical learning. 
     model ((D) Hierarchical) shows the best performance among the design choices.%\djkim{(another design choice ``TwoStream'' added.)}
	%, (D):Hierarchical target+Net.
	}
    \resizebox{.8\linewidth}{!}{
        \begin{tabular}{l| ccc}\hline
		&		Full&Rare&Non-rare\\\hline\hline
	%	(A) Modified Baseline  &   19.19	& 13.40  & 21.20\\
		(A) Modified Baseline &   19.09	& 13.09  & 20.89\\
		(B) MultiTask &   19.54	&13.93   &21.22\\
		(C) TwoStream &   19.63	& 13.67   & 21.41\\
		(D) Hierarchical &   \textbf{20.03}	&\textbf{14.52}   &\textbf{21.67}\\
	%	(D)  &   19.99	& 14.26   &21.71\\
		\hline
	\end{tabular}
	}
\label{table:architecture}	
\end{table}

V-COCO (Verbs in COCO) is a subset of MS-COCO~\cite{lin2014microsoft}, which consists of 10,346 images (2,533, 2,867, 4,946 for training, validation and test, respectively) and 16,199 human instances. 
Each person is annotated with binary labels for 26 action classes.
For the evaluation metric, we use the AP score as done for the evaluation on HICO-Det.
%\djkim{(maybe remove this sentence?$\xrightarrow{}$)} \steve{agreed} Compared to the HICO-Det datset, the V-COCO dataset has much fewer HOI classes and less diversity, thus we mainly evaluate our model on HICO-Det dataset.

\begin{table*}[t]
\centering
    \begin{minipage}{0.63\textwidth}
\caption{Results on the HICO-Det dataset compared to the existing state-of-the-art methods. 
%$\dagger$ indicates the usage of a more powerful object detector. 
The first row section shows the scores reported in the previous works while the second row section shows the results of our implementations.
The performance of our method consistently surpasses those of the existing HOI detectors.%Our model shows favorable performance against the current state-of-the-art models on all the metrics.%\djkim{+ The ``Known Object'' score of our model still outperforms the state-of-the-art models by a large margin.}
}
    \resizebox{1\linewidth}{!}{
        \begin{tabular}{l| ccc|ccc}\hline
			&\multicolumn{3}{c|}{Default}&\multicolumn{3}{c}{Known Object}\\
			&		Full&Rare&Non-rare &Full&Rare&Non-rare \\\hline\hline
			Shen~\etal~\cite{shen2018scaling}	            & 6.46	& 4.24  & 7.12 &-&-&-\\
			HO-RCNN~\cite{chao2018learning}                 & 7.81	& 5.37  & 8.54 & 10.41 & 8.94 & 10.85\\
			Gupta~\etal~\cite{gupta2015visual} impl. by~\cite{gkioxari2018detecting}  &   9.09	&7.02   &9.71&-&-&-\\
			InteractNet~\cite{gkioxari2018detecting}	    & 9.94	& 7.16  & 10.77 &-&-&-\\
			GPNN~\cite{qi2018learning}	                    & 13.11 & 9.34  & 14.23 &-&-&-\\
			iCAN~\cite{gao2018ican}	                        & 14.84 & 10.45 & 16.15 & 16.43& 12.01 & 17.75 \\
			%iHOI~\cite{xu2019interact}                      & 13.39 & 9.51  & 14.55 &-&-&-\\
			%with Knowledge~\cite{xu2019learning}	        & 14.70 & 13.26 & 15.13 &-&-&-\\
			Interactiveness Prior~\cite{li2019transferable} & 17.22 & 13.51 & 18.32 & 19.17 & 15.51 & 20.26\\
			Contextual Attention~\cite{wang2019deep}        & 16.24 & 11.16 & 17.75 & 17.73 & 12.78 & 19.21\\
			No-Frills~\cite{gupta2019no}	                & 17.18 & 12.17 & 18.68 &-&-&-\\
			RPNN~\cite{zhou2019relation}                    & 17.35 & 12.78 & 18.71 &-&-&-\\
			PMFNet~\cite{wan2019pose}                       & 17.46 & 15.65 & 18.00 & 20.34 & 17.47 & 21.20\\
			Peyre~\etal~\cite{peyre2019detecting}	        & 19.40 & 15.40 & 20.75 &-&-&-\\
			
			%Interaction Points~\cite{wang2020learning}            &   19.56	& 12.79 & 21.58 &22.05&15.77&23.92\\
			%VGSNet~\cite{ulutan2020vsgnet}                        &   19.80	& 16.05 & 20.91 &-&-&-\\
			%PPDM-DLA~\cite{liao2020ppdm}                          &   19.02	& 12.65 & 20.92 &-&-&-\\
			%PPDM-Hourglass~\cite{liao2020ppdm}                    &   21.10	& 14.46 & 23.09 &-&-&-\\
			PPDM~\cite{liao2020ppdm}                    &   21.10	& 14.46 & 23.09 &-&-&-\\
			DJ-RN~\cite{li2020detailed}                           &   21.34	& 18.53 & 22.18 &23.69&20.64&24.60\\
			
			%Heterogeneous~\cite{wangcontextual}                          &   17.57	& 16.85 & 17.78 &21.00&20.74&21.08\\
			%UnionDet~\cite{kim2020UnionDet}                               &   17.58	& 11.72 & 19.33 &19.76&14.68&21.27\\
			FCMNet~\cite{liuamplifying}                                 &   20.41	& 17.34 & 21.56 &22.56&18.97&23.12\\
			%PD-Net~\cite{zhong2020polysemy}                                 &   19.99	& 14.95 & 21.50 &24.15&18.06&25.97\\
			%PD-Net~\cite{zhong2020polysemy}$^\dagger$                        &   20.81	& 15.90 & 22.28 &24.78&18.88&26.54\\% with fusion with
			VCL~\cite{hou2020visual}                                   &   19.43	& 16.55 & 20.29 &22.00&19.09&22.87\\
			%VCL~\cite{hou2020visual}$^\dagger$                          &   23.63	& 17.21 & 25.55 &25.98&19.12&28.03\\% with finetune
			DRG~\cite{gao2020drg}                                   &   19.26	& 17.74 & 19.71 &23.40&21.75&23.89\\
			%DRG~\cite{gao2020drg}$^\dagger$                          &   24.53	& 19.47 & 26.04 &27.98&23.11&29.43\\% with finetune
			\hline
			\hline
			No-Frills (Reproduced)	  &   17.56	&13.23   &18.85 &22.02&16.97&23.53\\
			No-Frills + \textbf{ACP (Ours)}	  &   {20.59}	& \textbf{15.92}   & {21.98} & {25.35} & \textbf{19.55} & {27.08}\\
			No-Frills + \textbf{ACP++ (Ours)}	  &  \textbf{21.27}&{15.41}&\textbf{23.02} & \textbf{25.61} & {18.93} & \textbf{27.60}\\
			\hline
			PPDM (Reproduced)	  &   20.81	&13.69   &22.94 &-&-&-\\
			PPDM + \textbf{ACP++ (Ours)}	  &   \textbf{22.11}	& \textbf{14.43}   & \textbf{24.40} & - & - & -\\
			\hline
			DRG (Reproduced)	  &   18.77	& 16.41   &19.47 &24.74&23.74& 25.04 \\
			DRG + \textbf{ACP++ (Ours)}	  &   \textbf{18.90}	& \textbf{16.80}   & \textbf{19.52} & \textbf{24.78} & \textbf{23.87} & \textbf{25.05}\\
			\hline
		\end{tabular}
	}
	\label{table:hico}
	\end{minipage}
	\hfill
	\begin{minipage}{0.33\textwidth}
	\caption{Results on the V-COCO dataset. %$\dagger$ indicates the usage of a more powerful object detector. 
	{For our method, we show results both for constructing ACP from V-COCO and for using ACP constructed from HICO-Det instead. Both of these models show favorable performance against the current state-of-the-art models.}}
    \resizebox{1\linewidth}{!}{
        \begin{tabular}{l| c}\hline
    			&		$AP_{role}$\\%&$AP_{agent}$\\
    			\hline\hline
    			Gupta~\etal~\cite{gupta2015visual} impl. by~\cite{gkioxari2018detecting}	  &   31.8	\\%&65.1\\
    			InteractNet~\cite{gkioxari2018detecting}	  &   40.0	\\%&69.2 \\
    			GPNN~\cite{qi2018learning}	  &   44.0\\%&-\\
    			iCAN~\cite{gao2018ican}	  &   45.3\\%&-\\
    			%Turbo Learning	  &   42.0&70.3\\
    			iHOI~\cite{xu2019interact} & 45.79 \\%& -\\
    			with Knowledge~\cite{xu2019learning}	  &   45.9\\%&-\\
    			Interactiveness Prior~\cite{li2019transferable}	  &   48.7\\%&-\\
    			
    			Contextual Attention~\cite{wang2019deep} & 47.3\\%&- \\
    			RPNN~\cite{zhou2019relation}&47.53\\%&-\\
    			PMFNet~\cite{wan2019pose} & 52.0 \\%&-\\
    			
    			VCL~\cite{hou2020visual} & 48.3\\
    			DRG~\cite{gao2020drg} & 51.0\\
    			VSGNet~\cite{ulutan2020vsgnet} & 51.76 \\
    			Interaction Points~\cite{wang2020learning} & 52.3 \\
    			
    			%PD-Net~\cite{zhong2020polysemy} & 51.6 \\
    			PD-Net~\cite{zhong2020polysemy} & 52.6 \\% $^\dagger$
    			Heterogeneous~\cite{wangcontextual} & 52.7 \\
    			FCMNet~\cite{liuamplifying} & 53.1 \\
    			UnionDet~\cite{kim2020UnionDet} & 47.5 \\
    			%UnionDet~\cite{kim2020UnionDet}?? & 56.2 \\
    			\hline
    			%\textbf{ACP (Ours)}	  &   \textbf{52.64}	\\%&\textbf{?} \\
    			Our baseline            &       48.91   \\
    			\textbf{ACP (Ours, V-COCO)}	  &   \textbf{52.98}	\\%&\textbf{?} \\
    			\textbf{ACP (Ours, HICO-Det)}	  &   \textbf{53.23}	\\
    			\hline
    	\end{tabular}
    	}
    \label{table:vcoco}
	\end{minipage}
	
\end{table*}

\subsection{Quantitative Results}
\noindent\textbf{Ablation study}
In the ablations, the `No-Frills' baseline network~\cite{gupta2019no} we used is denoted as \emph{\textbf{Baseline}}. 
We first evaluate the effectiveness of the core design components in the proposed method, including (1) our simple modification to \emph{\textbf{Baseline}} in \Sref{sec.architecture}, denoted as \textbf{\emph{Modified Baseline}}; (2) the hierarchical learning technique introduced in \Sref{sec.architecture}, denoted as \emph{\textbf{Hierarchical}}; and (3) the knowledge distillation technique presented in {\Eref{eqn:total_loss}} of \Sref{sec.knowledge_distillation}, denoted as \emph{\textbf{Distillation}}.
Moreover, we show additional baselines which include either the self-attention module from \Sref{sec.attention} (denoted as \emph{\textbf{SA}}) or the word embedding loss function from \Sref{sec.language} (denoted as \emph{\textbf{$\mathcal{L}_{emb}$}}).  %, which we denote as \emph{\textbf{Extension (ACP++)}}.
\Tref{table:ablation} gives a comprehensive evaluation for each component. We draw conclusions from it one-by-one.

\emph{\textbf{First, our baseline network is strong.}} Our \emph{\textbf{Modified Baseline}} achieves 19.09 mAP and surpasses the `No-Frills' \emph{\textbf{Baseline}} by 1.51 mAP (a relative 8.7\% improvement). This is already competitive to the state-of-the-art result~\cite{peyre2019detecting} and serves as a strong baseline. 

\emph{\textbf{Second, both hierarchical learning and knowledge distillation are effective.}} This is concluded by adding \emph{\textbf{Hierarchical}} and \emph{\textbf{Distillation}} to the \emph{\textbf{Modified Baseline}}, respectively. Specifically, \emph{\textbf{+Hierarchical}} improves the modified baseline by 0.94 mAP (a relative 4.9\% improvement), and \emph{\textbf{+Distillation}} {({training with \Eref{eqn:total_loss}})} improves the modified baseline by 0.89 mAP (a relative 4.7\% improvement). Including both obtains 1.16 mAP improvement (relatively better by 6.1\%).

\emph{\textbf{Third, the proposed ACP method achieves a new state-of-the-art.}} Our final result is generated by further using the \emph{\textbf{PostProcess}} step (introduced in \Sref{sec.knowledge_distillation}) that projects the final action prediction into the ACP constrained space. Our method achieves 20.59 mAP (relative 7.9\% improvement) for Full HOI categories, 15.92 mAP (relative 21.6\% improvement) for Rare HOI categories, and 21.98 mAP (relative 5.2\% improvement) for Non-rare HOI categories. 
Note that our method made especially significant improvements for Rare classes, which supports the claim that the proposed method can alleviate the long-tailed distribution problem of HOI detection datasets.
This result set the new state-of-the-art at the time of submission for our earlier version of this work~\cite{kim2020detecting} on both the HICO-Det and V-COCO datasets as shown in \Tref{table:hico} and \Tref{table:vcoco}.

\emph{\textbf{Fourth, the extended ACP++ outperforms our previous ACP.}}
The word embedding loss ($\mathcal{L}_{emb}$) improves the performance of our previous ACP model in all the metrics, especially on rare classes, by alleviating the bias in the dataset.
In addition, self-attention module (SA) enhances the representation power of the model, thus it elevates the overall mAP by improving the performance on non-rare classes. 
However, SA reduces the performance on rare classes.
We conjecture that the model overfits on the major classes due to the additional parameters introduced by the SA module and degrades the generalization ability on rare classes.
%By combining both of the technical extensions, the overall performance further improves.}
Combining both of the technical extensions upon our previous ACP model gives the final extended ACP++ result.
%Our extended result is achieved by further adding \emph{\textbf{+Extension}} to our previous ACP model.
Without \emph{\textbf{PostProcess}}, our extended method improves upon our original ACP method by 20.96 mAP (relative 3.5\% improvement), and our extended method improves upon our original ACP method by 21.27 mAP (relative 3.3\% improvement) with \emph{\textbf{PostProcess}}.

In addition, the \textbf{\emph{MultiTask}}, \textbf{\emph{TwoStream}}, and our \textbf{\emph{Hierarchical}} architecture are compared in \Tref{table:architecture}. From \textbf{\emph{MultiTask}}, it can be seen that the \emph{softmax} based anchor action classification already brings benefits to the \emph{\textbf{Modified Baseline}} when used only in a multi-task learning manner. 
{From \textbf{\emph{TwoStream}}, separately modeling the anchor and the regular classes leads to slightly more improvement compared to \textbf{\emph{MultiTask}}}.
Moreover, our \textbf{\emph{Hierarchical}} architecture  improves upon \textbf{\emph{TwoStream}} by explicitly modeling the hierarchy between anchor and regular action predictions.

\begin{table}[t]
\centering
    \caption{Comparison of our model to the Functional Generalization method of Bansal~\etal~\cite{bansal2020detecting}. 
	%The first row indicates the final result reported by \cite{bansal2020detecting} including elaborate tuning (+engineering) that was not explained in detail and cannot be reproduced.
	The top set of rows shows the scores of the baseline and the Functional Generalization model without tuning as reported in \cite{bansal2020detecting}.
	In bottom set of rows, we show the results of our implementations that include the Functional Generalization model and our ACP method.
	The ACP method can achieve larger performance gains than Functional Generalization.
	%The model with both functional generalization and our ACP method shows the best performance with a large performance gap.
	}
    \resizebox{1\linewidth}{!}{%
		\begin{tabular}{l| ccc}\hline
			&		Full&Rare&Non-rare\\\hline\hline
    		%Functional Generalization (+engineering)~\cite{bansal2020detecting} &   21.96	& 16.43  & 23.62\\\hline\hline
    		Baseline of \cite{bansal2020detecting} &   12.72	& 7.57  & 14.26\\
    		Functional Generalization~\cite{bansal2020detecting} &   14.35	& 9.84  & 15.69\\\hline
			Baseline of \cite{bansal2020detecting} (Reproduced) &   13.09	&7.87   &14.65\\
			Functional Generalization (Reproduced) &   13.68	& 8.62   &15.19\\
			Functional Generalization + ACP (\textbf{Ours}) &   \textbf{15.38}	& \textbf{10.30} & \textbf{16.90}\\
			\hline
		\end{tabular}
    }
	\label{table:funcgen}	
\end{table}

\noindent\textbf{Comparison with the state-of-the-art}
We compare our method with the previous state-of-the-art techniques in \Tref{table:hico}.
The first and second column sections in \Tref{table:hico} are for `Default' and `Known Object' matrices, respectively.
Among the methods included in this comparison are the benchmark model of the HICO-Det dataset~\cite{chao2018learning}, the baseline models that we trained or modified from~\cite{gao2020drg,gupta2019no,liao2020ppdm}, and the current published state-of-the-art methods. 
As shown in \Tref{table:hico}, our ACP model (Ours) shows consistent improvements over our baseline models on all metrics, with especially significant gains over the `No-Frills' baseline, and shows favorable performance against the current state-of-the-art models in terms of all the metrics. 
In particular, our model (No-Frills + ACP++) surpasses the state-of-the-art model at the time of submission~\cite{peyre2019detecting} by 1.87 mAP.% for Full HOI categories.%, 0.52 mAP for Rare HOI categories, and 1.23 mAP for Non-rare HOI categories. 
We also added the result of applying ACP++ priors to more recent HOI detection methods~\cite{gao2020drg,liao2020ppdm}, and this also achieves consistent performance improvements for all the metrics. Note that `PPDM + ACP++' model outperforms the state-of-the-art model without elaborate fine-tuning~\cite{li2020detailed}.

\noindent\textbf{Results on V-COCO dataset}
To show the generalization ability of our method, we also evaluate it on the V-COCO dataset.
Note that the exact same method is directly applied to both HICO-Det and V-COCO, including the co-occurrence matrix, anchor action selection, and architecture design.
{We also constructed a co-occurrence matrix from V-COCO, but the matrix was sparse. Thus, to better take advantage of our idea, we instead use the co-occurrence matrix collected from the HICO-Det dataset to train on V-COCO.}
\Tref{table:vcoco} shows the performance of our model implemented upon the `No-Frills' model (Ours, HICO-Det) compared to the recent state-of-the-art HOI detectors on the V-COCO dataset.
{In addition, we show results of our model with the co-occurrence matrix constructed from the V-COCO dataset (Ours, V-COCO).}
{Both of these models show} favorable performance on the V-COCO dataset against the existing state-of-the-art models~\cite{liuamplifying}.
%We conclude that our ACP method is easily generalizable to other datasets. %without any requirements or limitations.

\noindent\textbf{Experiments with Functional Generalization~\cite{bansal2020detecting}}
We also examine our approach together with an HOI detection method called `Functional Generalization'~\cite{bansal2020detecting}.
Since its paper does not provide the details for its elaborate fine-tuning (which led to improvements from 14.35 mAP to 21.96 mAP), we consider it to be unsuitable for comparison in \Tref{table:hico}. However, we found our method to be complementary to its untuned version.
%Instead, in order to compare our method with \cite{bansal2020detecting}, we tried to reproduce the model of Bansal~\etal (both the baseline and the functional generalization method), and we add our ACP method upon this model. 
As shown in \Tref{table:funcgen}, according to our implementation, the functional generalization approach improves upon its baseline by 0.59 mAP (a relative 4.51\% improvement), while adding our ACP method to the functional generalization model improves it by 1.70 mAP (a relative 12.43\% improvement).
This indicates that our ACP method can complement other methods such as Functional Generalization.
%We conclude that our ACP method can also be helpful for improving HOI detection performance upon functional generalization method.
In addition, the results suggest that the ACP method can achieve larger performance gains than Functional Generalization.
Finally, note that the Functional Generalization method requires word vectors pre-trained on an external dataset, whereas our ACP prior can be constructed from only the labels of our target dataset.

\begin{table}[t]
\centering
    \caption{Results on the zero-shot triplets of the HICO-Det dataset. 
    Our final model shows better performance than Peyre~\etal by a large margin.
    %Note that our 
    Our ACP model under the zero-shot setting even outperforms the supervised setting of our baseline.
    }
    \resizebox{.8\linewidth}{!}{%
		\begin{tabular}{l| c}\hline
			&		mAP\\\hline\hline
			Peyre~\etal~\cite{peyre2019detecting} Supervised & 33.7\\
			Peyre~\etal~\cite{peyre2019detecting} Zero-Shot &   24.1\\
			Peyre~\etal~\cite{peyre2019detecting} Zero-Shot with Aggregation &   28.6\\
			\hline
			Ours Supervised (Modified Baseline) & 33.27\\%33.36$\pm$0.74\\
			Ours Zero-Shot (Modified Baseline) &   20.34	\\
			\textbf{Ours Zero-Shot (ACP)}   &   \textbf{34.95}\\%\textbf{29.34$\pm$3.60}	\\
			\hline
			\textbf{Ours Zero-Shot (ACP++)}   &   \textbf{35.12}\\%\textbf{29.34$\pm$3.60}	\\
			\hline
		\end{tabular}
    }
	\label{table:zero-shot}	
\end{table}

\begin{figure*}[t]
	\centering
		%\resizebox{1\linewidth}{!}{
		\includegraphics[width=.9\linewidth,keepaspectratio]{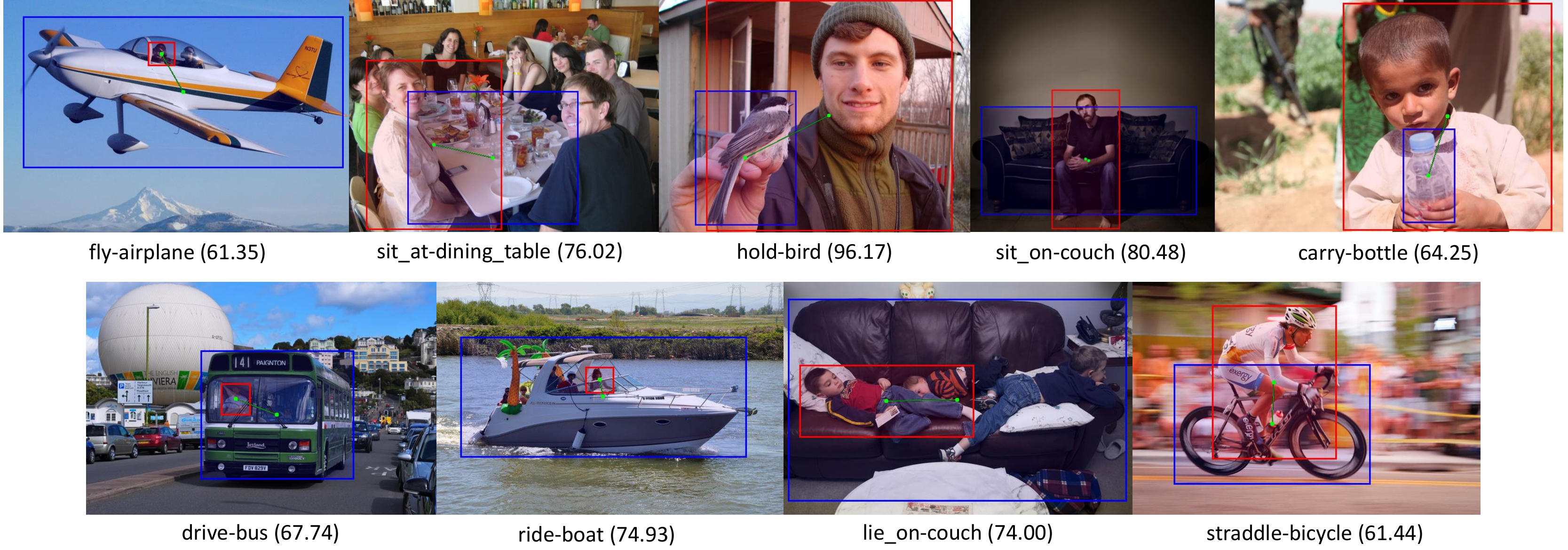}
		%\includegraphics[width=.9\linewidth,keepaspectratio]{Figures/qualitative2}
		%}
	\vspace{-2mm}
	\caption{Examples of bounding boxes and HOI detection scores from our model. Bounding boxes for humans are colored red, and bounding boxes for objects are colored blue. Each image is displayed with the predicted action+object class followed by the probability computed by our model.
	}
	\vspace{-2mm}
	\label{fig:qualitative}
\end{figure*}

\begin{figure}[t]
\centering
\includegraphics[width=.8\linewidth,keepaspectratio]{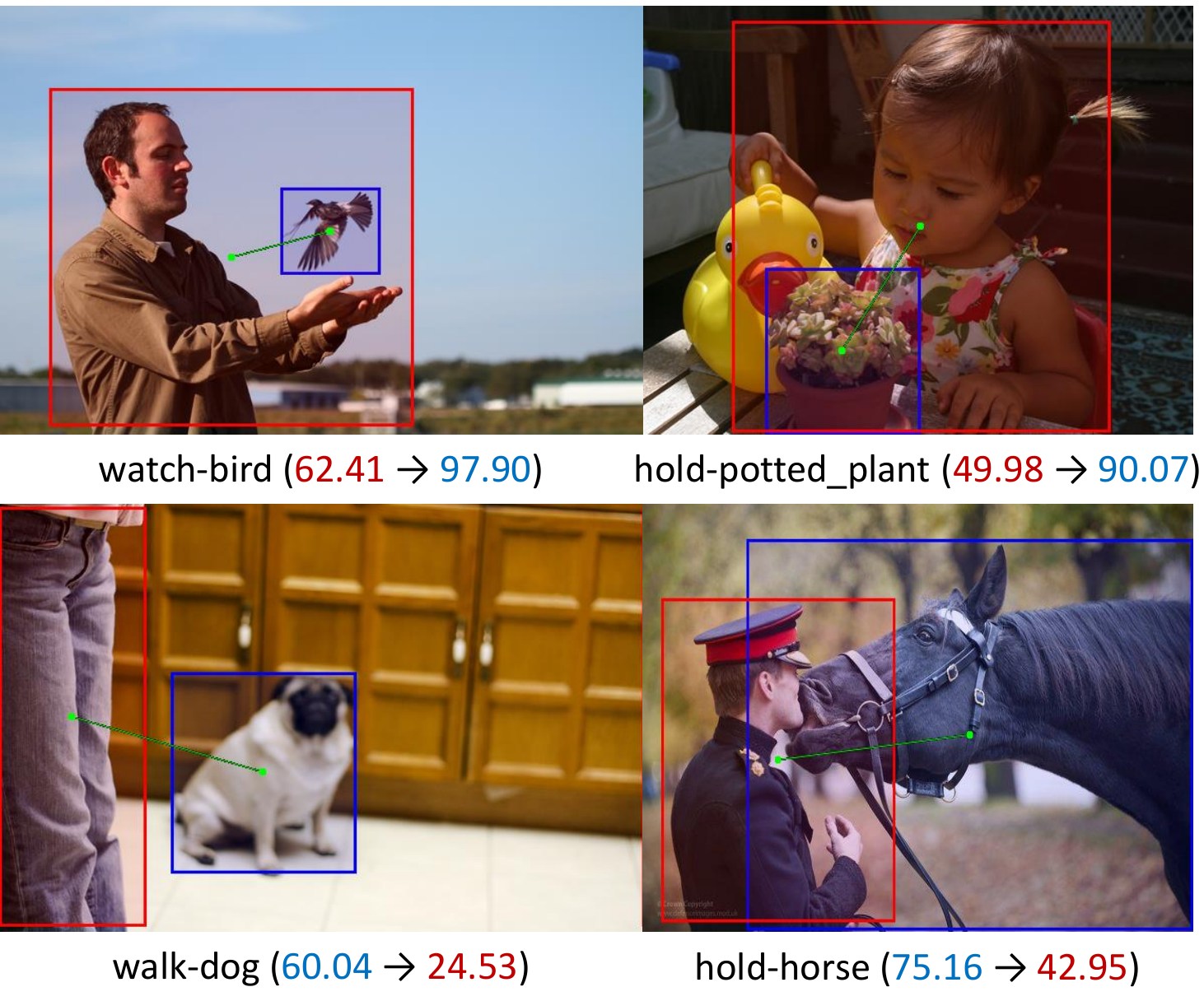}
\vspace{-2mm}
\caption{The HOI probability before and after applying the projection function $project(\cdot)$ on our model's HOI prediction (\emph{\textbf{PostProcess}}). 
%Leveraging co-occurrence matrix $C$ can not only increase the score for true classes (top) but also reduce the score for false classes (bottom). 
Note that \emph{\textbf{PostProcess}} can be done without any optimization.
}
\vspace{-2mm}
\label{fig:prob_change}
\end{figure}%

\noindent\textbf{Results of the zero-shot setup on the HICO-Det dataset}
The \emph{zero-shot setting} on the HICO-Det dataset was defined by Peyre~\etal~\cite{peyre2019detecting}. 
Specifically, we select a set of 25 HOI classes that we treat as unseen classes and exclude them and their labels in the training phase. 
However, we still let the model predict those 25 unseen classes in the test phase, which is known as the zero-shot problem. 
%Refer to Peyre~\etal~\cite{peyre2019detecting} for more information.
These HOI classes are randomly selected from among the set of non-rare HOI classes.
Since Peyre~\etal did not provide which specific HOI classes they selected, we select the unseen HOI classes such that the performance (mAP) for these classes in our \textbf{\emph{Modified Baseline}} model (introduced in \Sref{sec.architecture}) is similar to the corresponding \emph{\textbf{Supervised}} baseline in~\cite{peyre2019detecting}. 
In \Tref{table:zero-shot}, we show results of our final model (ACP and ACP++) and our modified baseline model compared to the corresponding setting reported in~\cite{peyre2019detecting}.
Our ACP model shows better performance (35.0 mAP) than Peyre~\etal (28.6 mAP) by a large margin (relative 22.4\% improvement).
In addition, our extended model (ACP++) further improves the performance of our ACP (35.1 mAP).
This result is remarkable in that our ACP model under the zero-shot setting even \emph{outperforms} the supervised setting of our baseline model, indicating the power of the proposed ACP method to effectively leverage prior knowledge on action co-occurrences. Furthermore, the analogy transfer method proposed by Peyre~\etal (denoted as aggregation) requires large-scale linguistic knowledge to train a word representation, whereas our model only requires the co-occurrence information of the labels in the dataset, which is much easier to obtain.
We conclude that the proposed method is effective for the zero-shot problem while being easy to implement.

\subsection{Qualitative Results and Analysis}
\begin{figure}[t]
\centering
\includegraphics[width=1\linewidth,keepaspectratio]{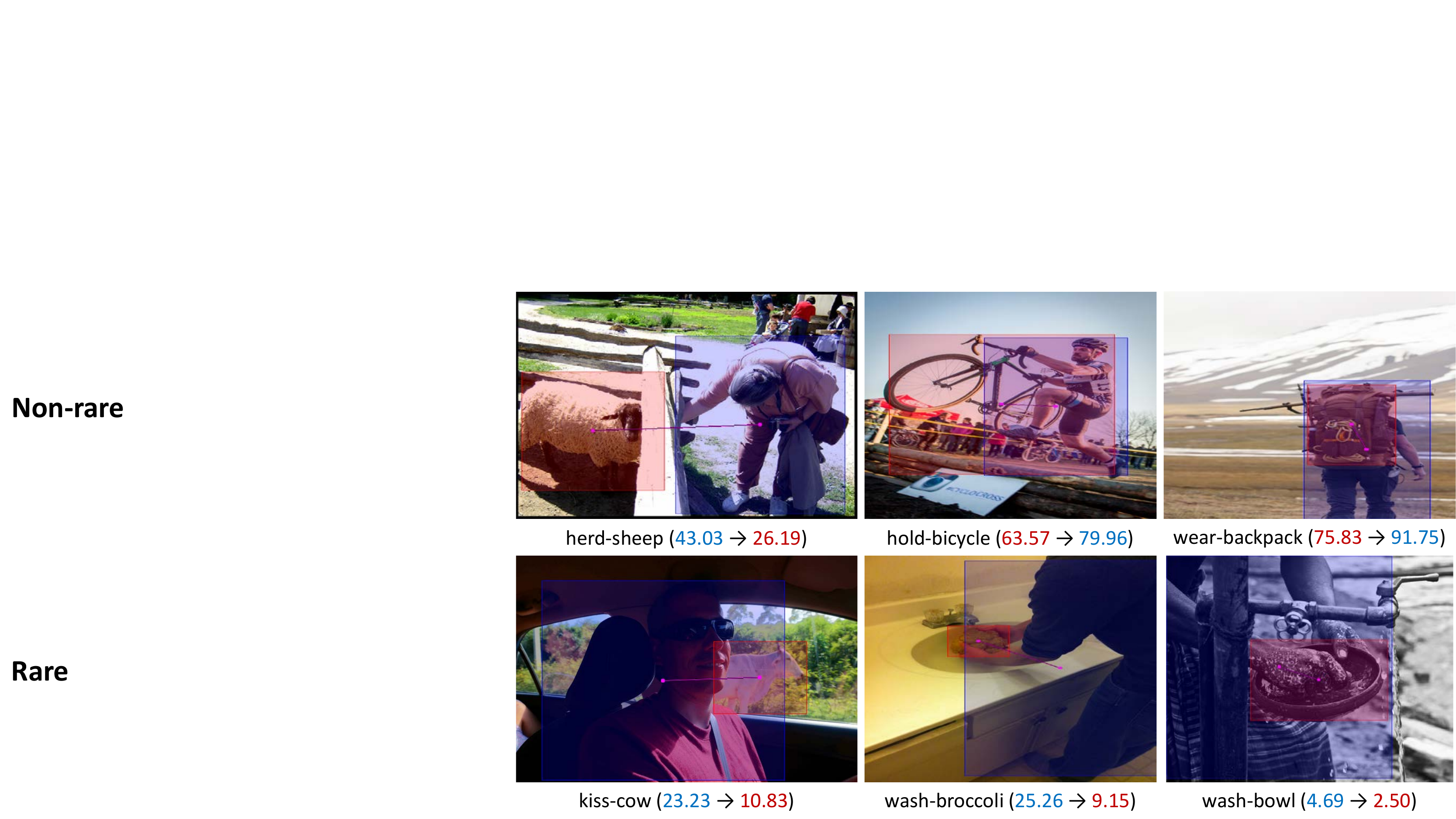}
\vspace{-4mm}
\caption{The HOI probability before and after applying the self-attention in our model for non-rare (top row) and rare (bottom row) HOI classes. For non-rare classes, our self-attention increases the score for true classes (`hold-bicycle' and `wear-backpack') and reduces the score for false classes (`herd-sheep'). On the other hand, for rare classes, the self-attention mostly reduces the scores even for true classes (\eg~`wash-bowl' and `wash-broccoli').
}
\vspace{-2mm}
\label{fig:self-attention}
\end{figure}

\noindent\textbf{Qualitative results}
In addition, \Fref{fig:qualitative} shows examples of HOI detection results that our model predicts correctly with high probability.
We show each image with the predicted HOI class followed by the probability computed by our model.
Also, we show the HOI probability change from before to after applying the projection function $project(\cdot)$ on our model's HOI prediction (\ie~the effect of \emph{\textbf{PostProcess}} introduced in \Sref{sec.knowledge_distillation}) in \Fref{fig:prob_change}.
Leveraging co-occurrence matrix $C$ not only increases the score for true classes (top) but also reduces the score for false classes (bottom). 
Note that this change can be achieved \emph{without any optimization} process.
Moreover, we show changes in HOI probability from before to after applying the self-attention in our model in \Fref{fig:self-attention}. It is found that our self-attention module mostly reduces the probability for rare classes (even for true classes), which might be the reason for the mAP degradation on rare classes shown in \Tref{table:ablation}.

\begin{figure}[t]
	\centering
		\includegraphics[width=.7\linewidth,keepaspectratio]{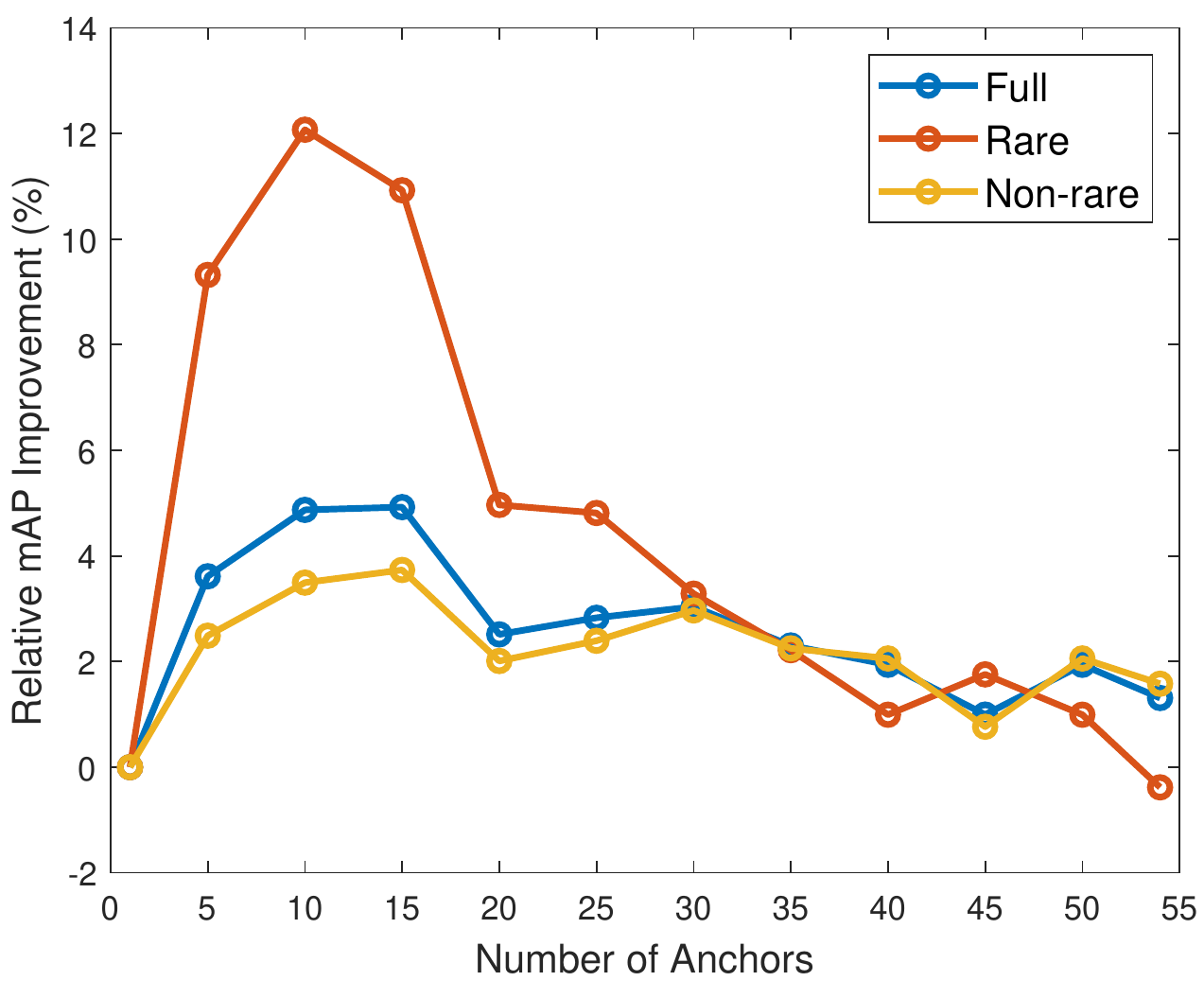}
	\vspace{-2mm}
	\caption{Performance of the hierarchical architecture with different numbers of anchors at intervals of five. The models with 15 and 10 anchors show the best performance overall and on rare classes, respectively.}
	\vspace{-2mm}
	\label{fig:num_of_group}
\end{figure}

\noindent\textbf{Number of anchor actions}
In addition, we investigate the effect of using different numbers of anchor actions $|\mathcal{D}|$ in \Fref{fig:num_of_group}. 
We measure the relative performance improvement from the \emph{\textbf{+Hierarchical}} model to the \emph{\textbf{Modified Baseline}} model by changing the number of anchor actions 
%$|\mathcal{D}|$ 
at intervals of five.

In principle, the more anchor actions we use, the better performance that can be attained. 
On one hand, the selected anchor actions can be more distinguishable from one another with more anchor action categories and training samples.
On the other hand, the remaining regular actions can also benefit from stronger co-occurrence priors. 
However, more anchor action will also result in more sub-networks to optimize, and this will cause over-fitting to a certain extent. 
Through observations, we found that an increase in the number of parameters of the HOI detector often causes a severe performance decrease.
Thus, there is a trade-off between a large and small number of anchors, which requires us to empirically select the best anchor number.
As depicted in \Fref{fig:num_of_group}, the hierarchical architecture shows the best overall mAP score (Full) with 15 anchors and the best mAP score on rare classes with 10 anchors.
We finally use an experimentally overall best-performing choice of 15 anchor actions (maximum anchor action number is 54).

%\noindent\textbf{Performance across different classes}
%\Fref{fig:class_score} shows the improvements in mAP scores across different object and action classes.
%We compare our model with the \emph{\textbf{Baseline}} model.%`No-Frills' model~\cite{gupta2019no}.%\emph{\textbf{Baseline}} model.
%In order to visualize on which classes our method improves performance, the figure is organized in the order of the improvements of the scores of each class.
%It is found that our method improves mAP scores in 64 out of the 80 object classes (80\%) and 90 out of the 117 actions (77\%).

%\begin{figure*}[t]
%	\centering
%		\includegraphics[width=1\linewidth,keepaspectratio]{Figures/score_per_obj3}\\
%		\includegraphics[width=1\linewidth,keepaspectratio]{Figures/score_per_verb3}\\
%	\caption{The AP score improvements across different object and action classes {sorted by the amount of improvement} compared to the baseline model~\cite{gupta2019no}. 
%	Our method improves mAP scores in 64 out of the 80 object classes (80\%) and 90 out of the 117 actions (77\%). 
%	}
%	\label{fig:class_score}
%\end{figure*}

\begin{figure}[b]
\centering
\includegraphics[width=.7\linewidth,keepaspectratio]{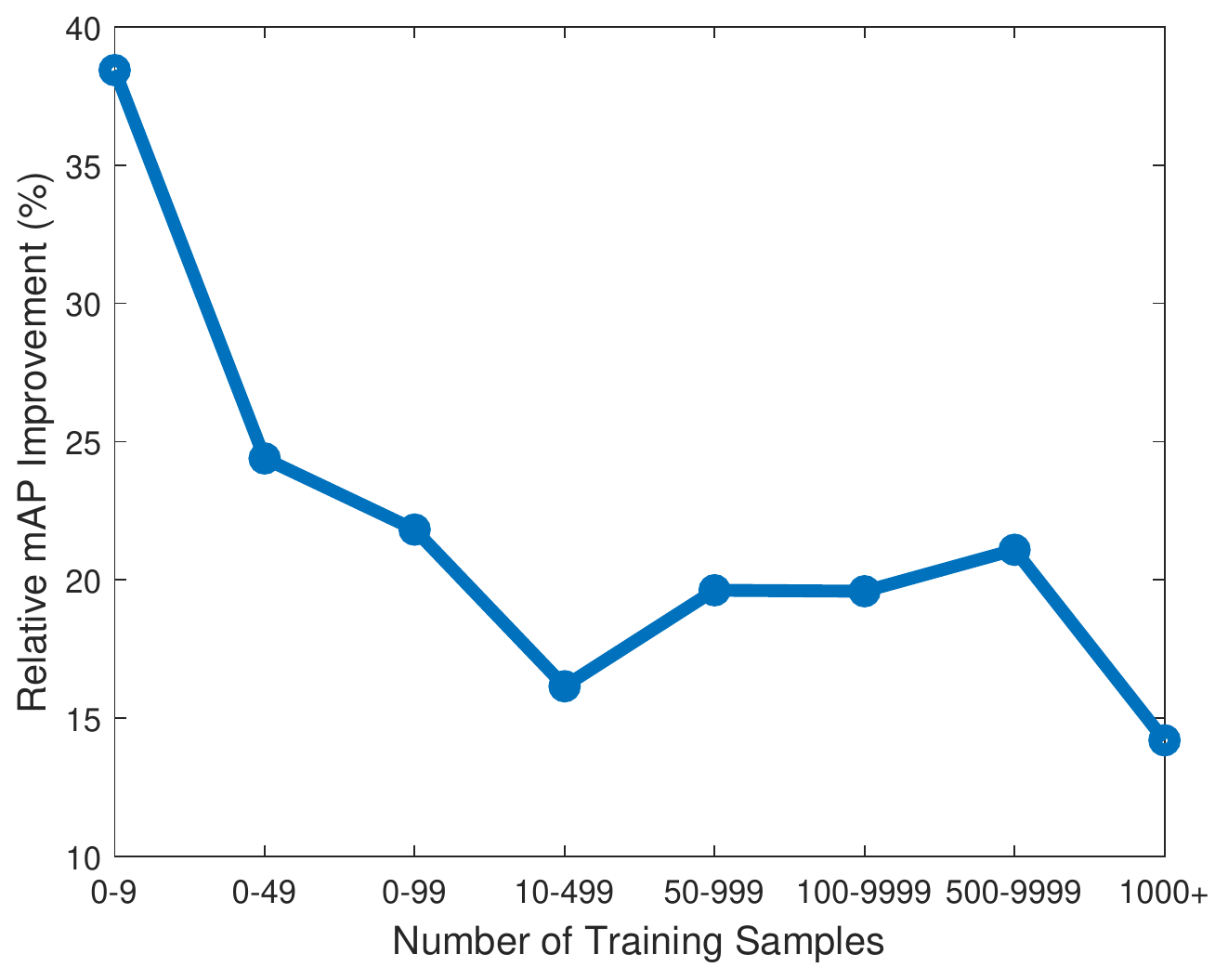}
\vspace{-2mm}
\caption{The relative mAP score improvements for various HOI sets with different numbers of training samples. Our method is able to improve the performance especially when the number of training samples is small (38.24\% improvement for 0-9 samples).
}
\vspace{-2mm}
\label{fig:num_of_sample}
\end{figure}

\noindent\textbf{Performance on various sets with different number of training samples}
{In \Fref{fig:num_of_sample}, we show the relative mAP score improvements of our model compared to the baseline model by computing mAP on various sets of HOI classes that have different number of training samples.
%As one can expect, for the HOI classes with larger number of training samples show higher mAP scores. 
Our method shows positive performance improvements for all numbers of training samples. 
Also, there is a trend that HOI classes with a small number of training samples mostly show larger performance improvements.
In particular, for HOI classes with the number of training samples between 0 and 9, our model achieves 38.24\% improvement compared to the baseline model.
These results indicate that the proposed method is able to improve the performance of an HOI detector, especially for classes with few training samples.
}

\begin{figure}[t]
	\centering
		\includegraphics[width=1\linewidth,keepaspectratio]{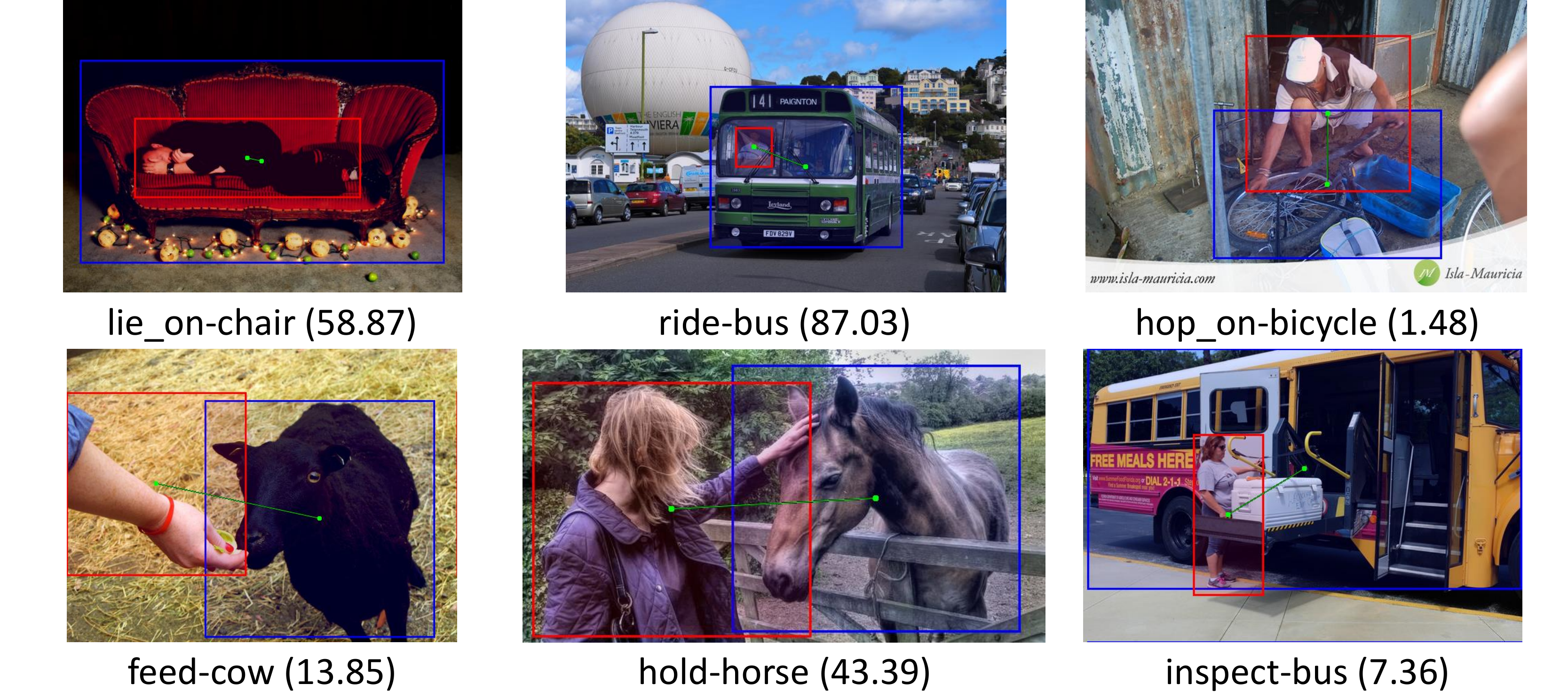}
	\vspace{-2mm}
	\caption{Examples of the false predictions from our model. The false cases include wrong prediction of object label from the object detector (left column), correct prediction but missing ground truth labels (middle column), and predictions that could have been correct if the context were taken into account (right column).}
	\vspace{-2mm}
	\label{fig:failure}
\end{figure}

\noindent\textbf{Analysis of false predictions}
\Fref{fig:failure} shows examples of false predictions by our model.
We found three common reasons for predictions being evaluated as false.
First is the wrong prediction of object label from the object detector (\eg~`couch' as `chair' or `sheep' as `cow') which is shown in the left column of \Fref{fig:failure}.
Second, the prediction is correct but the ground truth label for the prediction in a test image is missing (\eg~`ride-bus' or `hold-horse') which is shown in the middle column of \Fref{fig:failure}. 
Last, an HOI detector could have been predicted correctly if there were a sophisticated way to take context (the third object or the background) into account (\eg~the background for predicting `repair-bicycle' or an object that person carries for predicting `load-bus') which is shown in the right column of \Fref{fig:failure}.
The last issue could be solved by devising a better network architecture for effectively encoding context, which is a direction orthogonal to our work.
All of these issues (the errors in the object detector, missing labels in datasets, and encoding context) are fundamental issues in HOI detection, which can be interesting topics for future work.

\section{Conclusion}
We introduced a novel method to effectively train an HOI detector by leveraging prior knowledge on action co-occurrences in two different ways, via the architecture and via the loss function.
Our proposed method consistently achieves favorable performance compared to the current state-of-the-art methods in various setups.
%\steve{Our proposed method achieves favorable performance compared to the current state-of-the-art methods, but like these methods, it suffers from fundamental issues in HOI detection, such as wrong predictions from the object detector, missing ground truth labels, and not capturing global context.}
Co-occurrence information not only is helpful for alleviating the long-tailed distribution problem but also can be easily obtained. 
Given the co-occurrence action/interaction priors, one open question is how to expand the co-occurrence priors to a larger vocabulary and more general domains.
Therefore, a possible direction for future work would be obtaining more general co-occurrence priors by leveraging external knowledge from the web. Another direction for future work is to construct and utilize co-occurrence priors for other relationship-based vision tasks~\cite{johnson2017clevr,kim2019dense,kim2020dense,lu2016visual} or other problems related to the dataset bias~\cite{kim2020distribution,oh2021distribution,tang2020unbiased}.
%such as visual reasoning~\cite{johnson2017clevr}, visual relationship detection~\cite{lu2016visual}, and relational captioning~\cite{kim2019dense}.
%A direction for future work is to construct and utilize co-occurrence priors for other high-level vision tasks such as visual question answering and image captioning.
%djkim{Also, as fundamental issues in  HOI detection, we discuss several reasons for false predictions (including (1) wrong prediction from the object detector, (2) missing ground truth labels, and (3) global context) in the supplementary material.}
\\

\noindent{\textbf{Acknowledgements.}}
This work was supported by the Institute for Information \& Communications Technology Promotion (2017-0-01772) grant funded by the Korea government.

%\newpage
%\section{Supplementary}
%\input{5_sec_supplementary}

% use section* for acknowledgment
%\section*{Acknowledgment}
%The authors would like to thank...

% Can use something like this to put references on a page
% by themselves when using endfloat and the captionsoff option.
\ifCLASSOPTIONcaptionsoff
  \newpage
\fi

{\small
\bibliographystyle{ieee_fullname}
\bibliography{egbib}
}

% biography section
% 
% If you have an EPS/PDF photo (graphicx package needed) extra braces are
% needed around the contents of the optional argument to biography to prevent
% the LaTeX parser from getting confused when it sees the complicated
% \includegraphics command within an optional argument. (You could create
% your own custom macro containing the \includegraphics command to make things
% simpler here.)
%\begin{IEEEbiography}[{\includegraphics[width=1in,height=1.25in,clip,keepaspectratio]{mshell}}]{Michael Shell}
% or if you just want to reserve a space for a photo:

\begin{IEEEbiography}[{\includegraphics[width=1in,height=1.25in,clip,keepaspectratio]{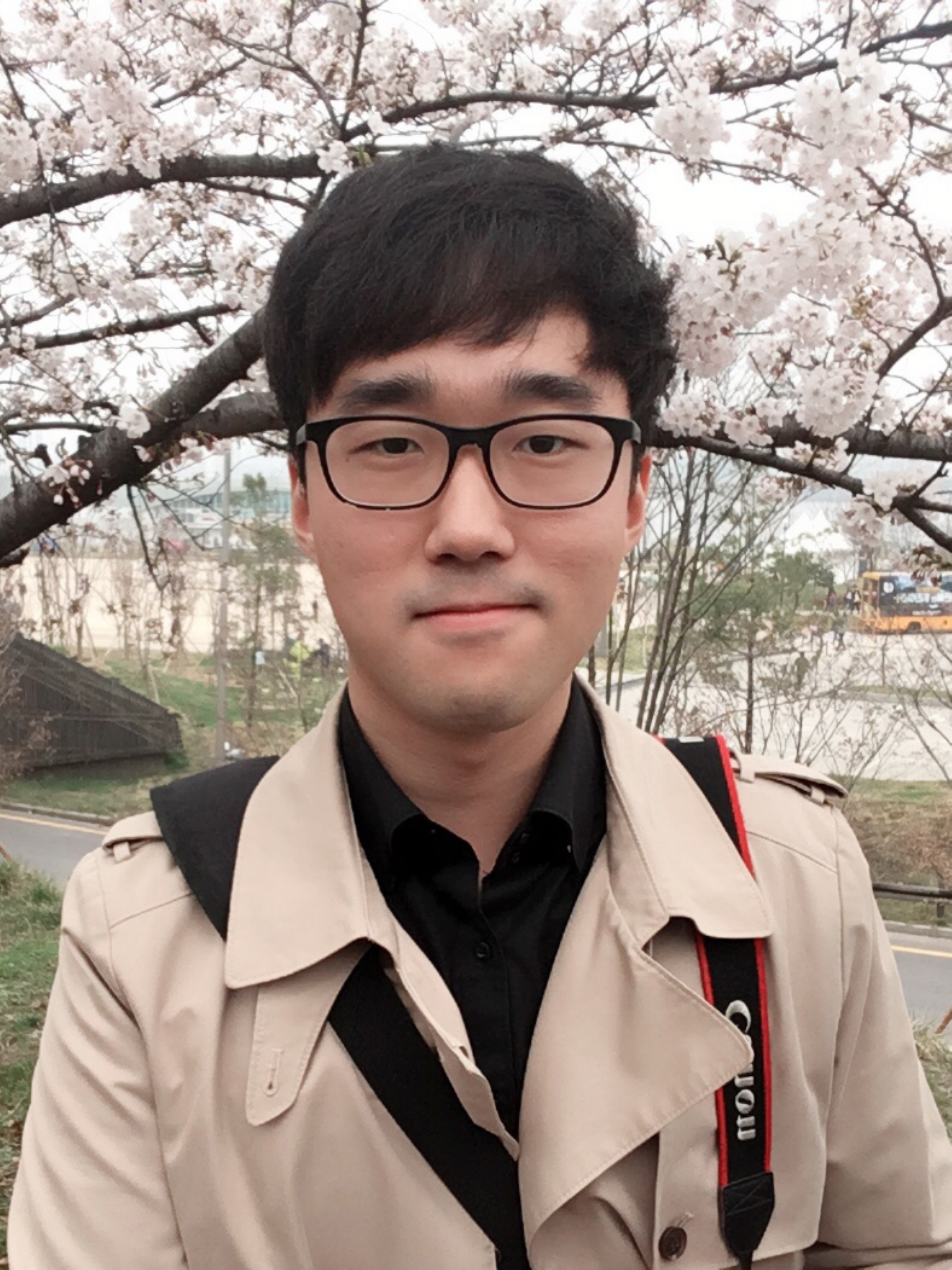}}]{Dong-Jin Kim}
received the B.S. degree, M.S. degree, and Ph.D. degree in Electrical Engineering from Korea Advanced Institute of Science and Technology (KAIST), Daejeon, South Korea, in 2015, 2017, and 2021, respectively.
%He is currently working towards the Ph.D. degree in Electrical Engineering at KAIST.
He was a research intern in the Visual Computing Group, Microsoft Research Asia (MSRA).
He was awarded a silver prize from Samsung Humantech paper awards and Qualcomm Innovation awards.
His research interests include high-level computer vision such as language and vision and human behavior understanding. 
He is a student member of the IEEE.
\end{IEEEbiography}

\begin{IEEEbiography}[{\includegraphics[width=1in,height=1.25in,clip,keepaspectratio]{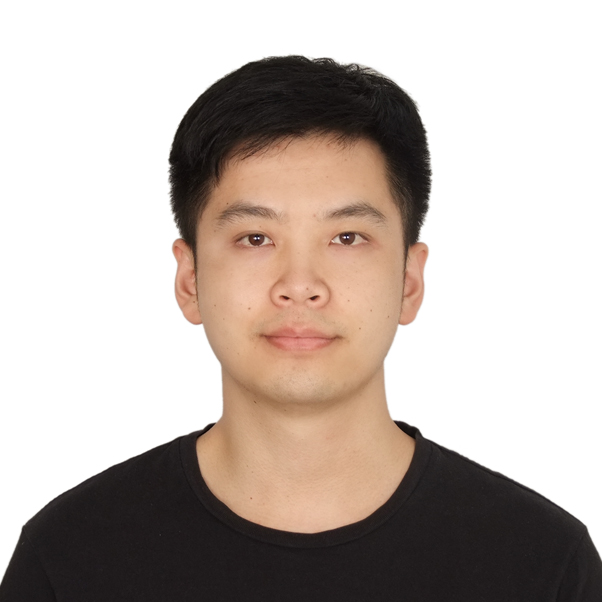}}]{Xiao Sun}
is a Senior Researcher in the Visual Computing Group of Microsoft Research Asia. He received the B.S. and M.S. degrees in Information Engineering from South China University of Technology, China, in 2011 and 2014, respectively. His research interests include computer vision and machine learning, especially pose estimation, object detection and action recognition.
\end{IEEEbiography}

\begin{IEEEbiography}[{\includegraphics[width=1in,height=1.25in,clip,keepaspectratio]{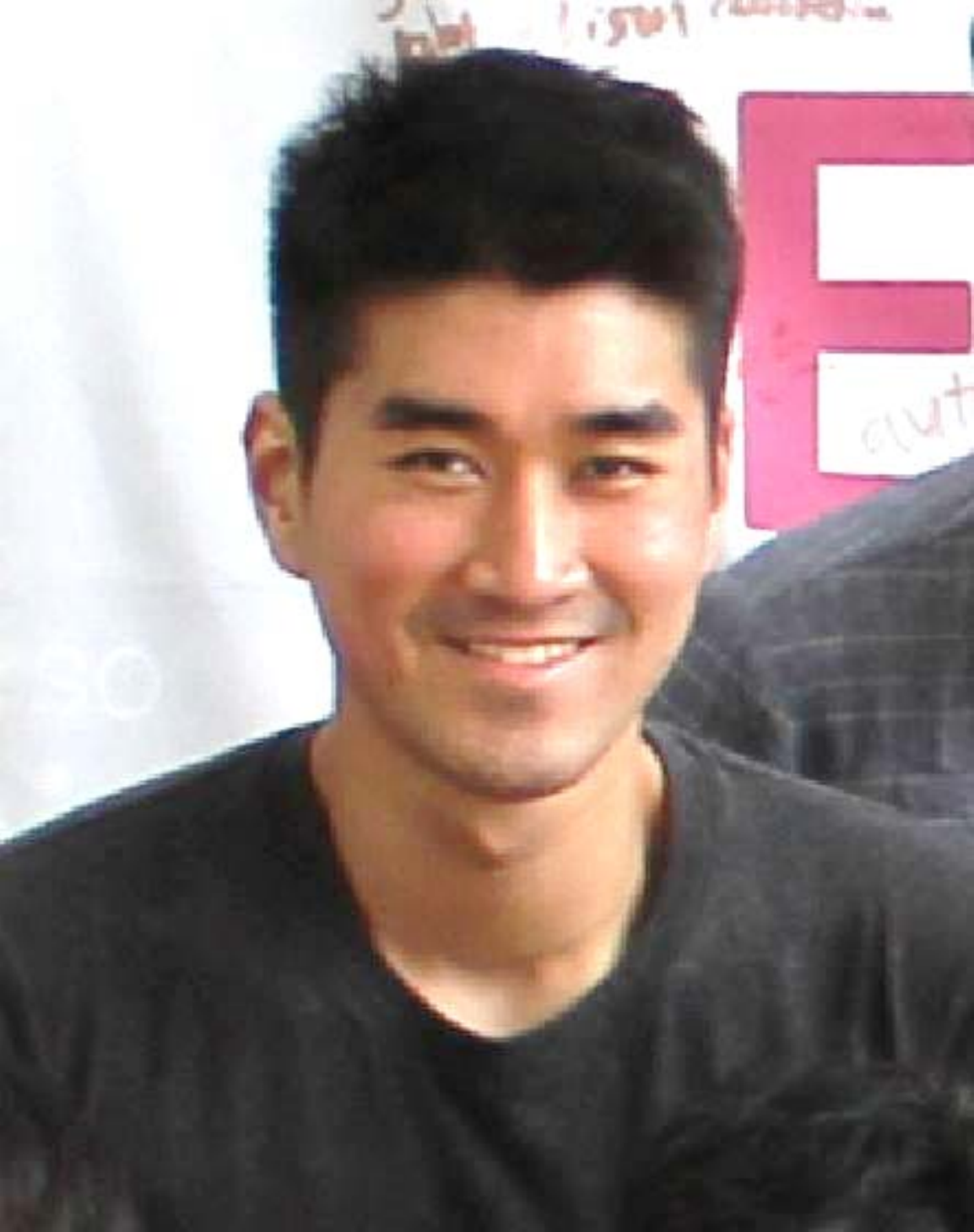}}]{Jinsoo Choi}
received his B.S., M.S., and Ph.D. degrees in Electrical Engineering from Korea Advanced Institute of Science and Technology (KAIST) in 2013, 2015, and 2020 respectively.
%He is an incoming machine learning video engineer at Apple.
He received the grand prize from the Electronic Times paper awards hosted by the Ministry of Science and ICT, Rep. of Korea, silver prize from Samsung Electro-Mechanics paper awards, silver prize from Samsung Humantech paper awards, Qualcomm  Innovation  awards, recognition as top research achievements and top 1\% research achievements from KAIST annual and biannual R\&D reports.
His research interests include deep learning, computer vision, and computer graphics with an emphasis on video enhancement and processing.
\end{IEEEbiography}

\begin{IEEEbiography}[{\includegraphics[width=1in,height=1.25in,clip,keepaspectratio]{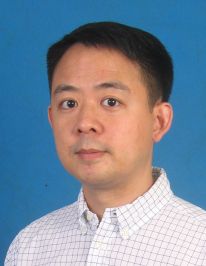}}]{Stephen Lin}
is a Senior Principal Research Manager in the Visual Computing Group of Microsoft Research Asia. 
He obtained a B.S.E. from Princeton University and a Ph.D. from the University of Michigan. His research interests include computer vision and computer graphics. 
Dr. Lin is on the editorial board of the International Journal of Computer Vision, and has served as a program chair for the International Conference on 3D Vision 2020, International Conference on Computer Vision 2011, and Pacific-Rim Symposium on Image and Video Technology 2009.
\end{IEEEbiography}

\begin{IEEEbiography}[{\includegraphics[width=1in,height=1.25in,clip,keepaspectratio]{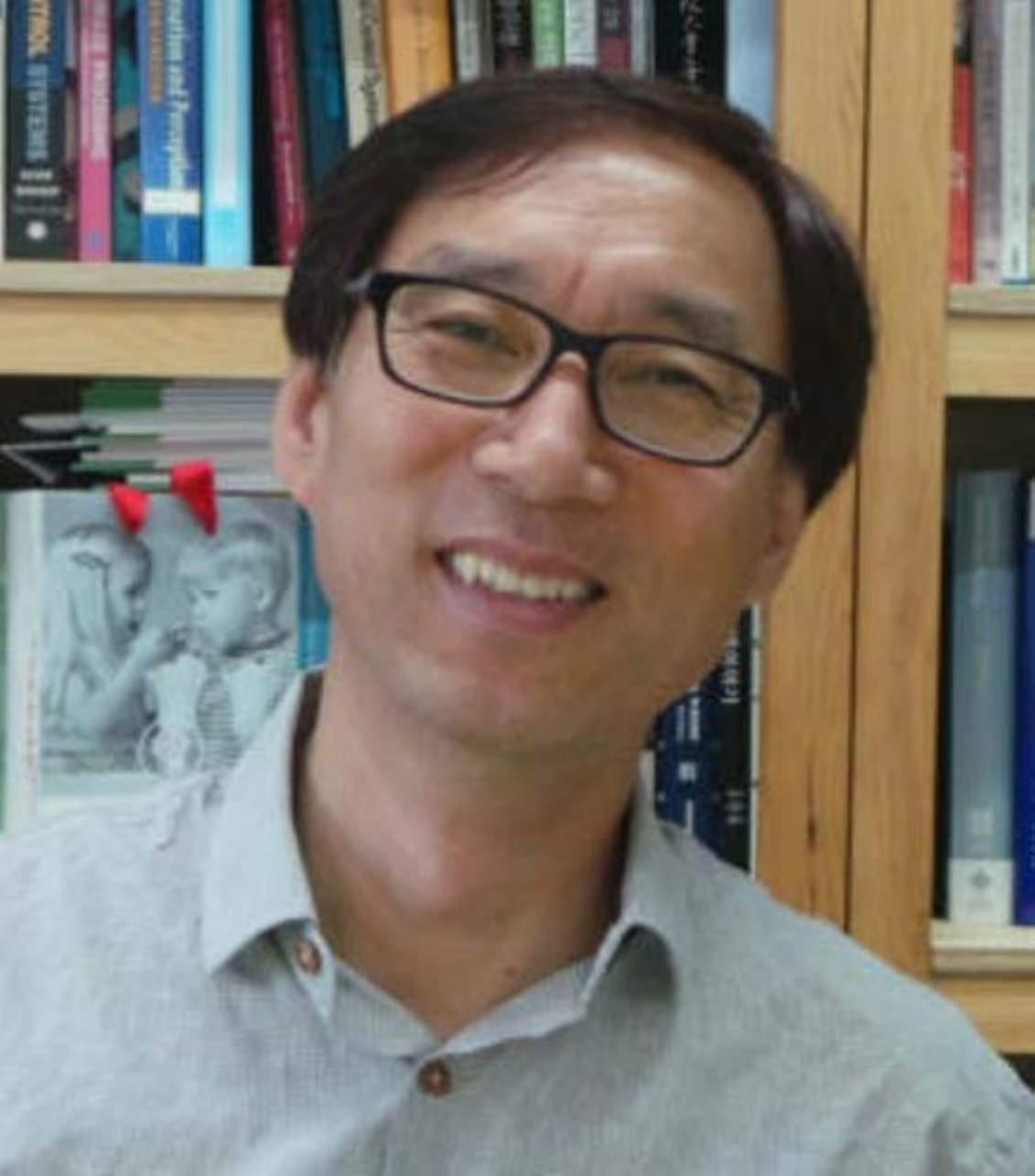}}]{In So Kweon}
received the BS and MS degrees in mechanical design and production engineering from Seoul National University, Seoul, South Korea, in 1981 and 1983, respectively, and the PhD degree in robotics from the Robotics Institute, Carnegie Mellon University, Pittsburgh, Pennsylvania, in 1990. He is an professor with the Electrical Engineering Department, KAIST, South Korea. He worked for the Toshiba R\&D Center, Japan, and joined the Department of Automation and Design Engineering, KAIST, Seoul, South Korea, in 1992, where he is currently a professor with the Department of Electrical Engineering. He is a recipient of the Best Student Paper Runner-up Award at the IEEE Conference on Computer Vision and Pattern Recognition (CVPR 09). His research interests include camera and 3D sensor fusion, color modeling and analysis, visual tracking, and visual SLAM. He was the program co-chair for the Asian Conference on Computer Vision (ACCV 07) and was the general chair for the ACCV 12. He is also on the editorial board of the International Journal of Computer Vision. He is a member of the IEEE
and KROS.
\end{IEEEbiography}

%\clearpage
%\section{contents for cover letter}
%\input{0_sec_cover_letter.tex}

\end{document}